\documentclass{article}
\PassOptionsToPackage{numbers, sort}{natbib}

\usepackage[final]{neurips_data_2023}

\usepackage[utf8]{inputenc} 
\usepackage[T1]{fontenc}    
\usepackage{hyperref}       
\usepackage{url}            
\usepackage{booktabs}       
\usepackage{amsfonts}       
\usepackage{nicefrac}       
\usepackage{microtype}      
\usepackage{xcolor}         
\usepackage{amsmath}
\usepackage{tabularx, makecell}
\usepackage{graphicx}
\usepackage{subfig}
\usepackage{multirow}
\usepackage{wrapfig}
\usepackage{longtable}

\newcolumntype{d}{l}

\newcolumntype{Y}{>{\centering\arraybackslash}X}
\newcolumntype{L}[1]{>{\raggedright\let\newline\\\arraybackslash\hspace{0pt}}m{#1}}
\newcolumntype{C}[1]{>{\centering\let\newline\\\arraybackslash\hspace{0pt}}m{#1}}
\newcolumntype{R}[1]{>{\raggedleft\let\newline\\\arraybackslash\hspace{0pt}}m{#1}}

\title{What a MESS: Multi-Domain Evaluation of\\Zero-Shot Semantic Segmentation}

\author{
  Benedikt Blumenstiel\thanks{Equal contributions}\\
  Karlsruhe Institute of Technology \\
  IBM Research Europe \\
  \texttt{benedikt.blumenstiel@kit.edu} \\
  \And
  Johannes Jakubik\textsuperscript{*}\\
  Karlsruhe Institute of Technology \\
  IBM Research Europe \\
  \texttt{johannes.jakubik@kit.edu} \\
  \And
  Hilde Kühne \\
  University of Bonn\\
  MIT-IBM Watson AI Lab\\
  \texttt{hildegard.kuehne@ibm.com} \\
  \And
  Michael Vössing \\
  Karlsruhe Institute of Technology \\
  IBM Germany \\
  \texttt{michael.voessing@kit.edu} \\
}

\begin{document}

\maketitle

\begin{abstract}
  While semantic segmentation has seen tremendous improvements in the past, there are still significant labeling efforts necessary and the problem of limited generalization to classes that have not been present during training. 
  To address this problem, zero-shot semantic segmentation makes use of large self-supervised vision-language models, allowing zero-shot transfer to unseen classes. 
  In this work, we build a benchmark for Multi-domain Evaluation of Semantic Segmentation (MESS), which allows a holistic analysis of performance across a wide range of domain-specific datasets such as medicine, engineering, earth monitoring, biology, and agriculture.
  To do this, we reviewed 120 datasets, developed a taxonomy, and classified the datasets according to the developed taxonomy. We select a representative subset consisting of 22 datasets and propose it as the MESS benchmark.
  We evaluate eight recently published models on the proposed MESS benchmark and analyze characteristics for the performance of zero-shot transfer models.
  The toolkit is available at \url{https://github.com/blumenstiel/MESS}.
\end{abstract}

\section{Introduction}

\begin{figure}[tbh]
    \centering
    \addtolength{\tabcolsep}{-0.5em}
    \begin{tabular}{lccccc}
 & FoodSeg103 & ISPRS Pots. & CHASE DB1 & DeepCrack & CUB-200 \\
\rotatebox[origin=l]{90}{Image} & 
\includegraphics[height=.15\textwidth]{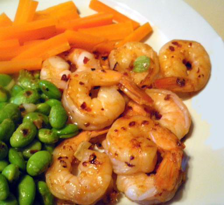} & 
\includegraphics[height=.15\textwidth]{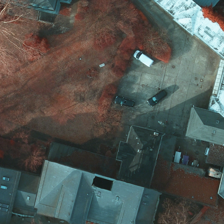} & 
\includegraphics[height=.15\textwidth]{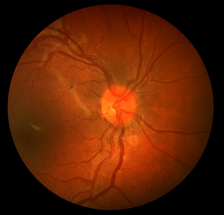} & 
\includegraphics[height=.15\textwidth]{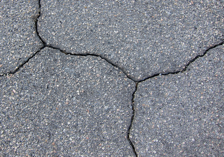} & 
\includegraphics[height=.15\textwidth]{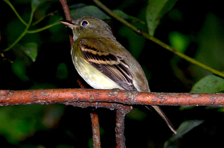} \\ 

\rotatebox[origin=l]{90}{Prediction} & 
\includegraphics[height=.15\textwidth]{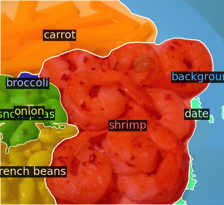} & 
\includegraphics[height=.15\textwidth]{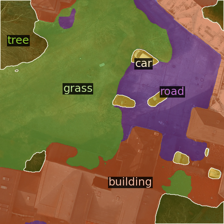} & 
\includegraphics[height=.15\textwidth]{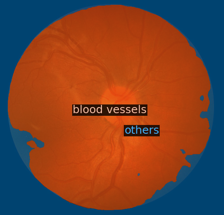} & 
\includegraphics[height=.15\textwidth]{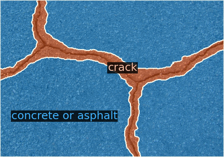} & 
\includegraphics[height=.15\textwidth]{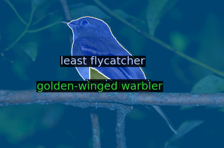} \\ 

\end{tabular}

    \caption{CAT-Seg-L~\cite{CAT-Seg2023} predictions for a range of domain-specific datasets. The model achieves promising predictions on everyday and satellite images, while it faces difficulties in segmenting small segments such as blood vessels, and distinguishing similar classes such as bird species.}
    \label{fig:catseg_examples}
\end{figure}

Zero-shot semantic segmentation utilizes self-supervised models such as CLIP to minimize labeling requirements during training and to improve model generalization. Recent models are already able to include classes during inference that were not present during training. 
For this reason, zero-shot semantic segmentation is becoming increasingly relevant for real-world scenarios.
In particular, the performance on domain-specific datasets such as earth monitoring datasets, as visualized in Figure~\ref{fig:catseg_examples}, becomes more and more relevant.
Current standard benchmarks tend to focus on in-domain tasks but do not capture performance comparisons across domains. 
This is problematic because it limits insight into the applicability of zero-shot semantic segmentation to new domains. It also makes it difficult to assess whether architectures might be suitable for datasets that pose additional challenges (e.g., different sensor types or specialized vocabulary).
To better understand the behavior of zero-shot semantic segmentation models on a wider range of more complex, domain-specific datasets, we propose a holistic Multi-domain Evaluation of Semantic Segmentation (MESS). 
To this end, we have examined 120 datasets and classified them within a developed taxonomy. 
We leverage our benchmark to evaluate eight recently published models for zero-shot semantic segmentation including the state-of-the-art models on 22 datasets from the fields of medical sciences, earth monitoring, agriculture and biology, engineering as well as a general domain including datasets on, e.g., driving scenes, maritime scenes, paintings, and body parts. 
Our evaluation focuses on zero-shot text-to-mask models---also known as open-vocabulary semantic segmentation (OVSS)---and later also compares their performance with zero-shot point-to-mask and box-to-mask approaches of SAM~\cite{SegmentAnything}.
Using the proposed benchmark, we identify and analyze several characteristics that influence the performance of OVSS models, i.a., showing that the semantic, textual similarity of classes as well as the underlying sensor type, significantly affect the performance of current models. 

Our experiments reveal various challenges for the application of zero-shot semantic segmentation on domain-specific datasets, e.g., we found that the selection of class labels can significantly affect the quality of predictions. We also observe that the models are sensitive to the semantics of the textual prompts, e.g., general terminology leads to better performance than domain-specific terminology. 
Overall, we hope that our benchmark will support accelerating zero-shot semantic segmentation and improve the real-world applicability of semantic segmentation in general.

We summarize the contributions of this work as follows: (1)~We develop a taxonomy based on a quantitative and qualitative analysis of a broad variety of semantic segmentation datasets. (2)~We propose a new benchmark for multi-domain semantic segmentation. (3)~We evaluate eight zero-shot models on the MESS benchmark with an in-depth analysis of the task characteristics.

\section{Related work}

\subsection{Zero-shot semantic segmentation}

Large-scale self-supervised pre-training has revolutionized the field of computer vision over the last couple of years. One stream of work focuses on vision-language pre-training such as in recent foundation model architectures like CLIP~\cite{CLIP2021}, ALIGN~\cite{ALIGN2021}, and Florence~\cite{Florence2021}. These models are trained on image-text pairs and encode both visual and text semantics in a shared embedding space. This approach particularly enables so-called open-vocabulary image classification by computing the similarity between the embeddings of the image and the embeddings of natural language describing the classes in the image. 
The text describing the images can be any arbitrary textual sequence and might describe classes on the images that have been unseen during training. This is in contrast to recent segmentation models, like Segment-Anything (SAM, see~\cite{SegmentAnything}), which are trained only on image data and therefore do not include a text encoder to encode semantic concepts. Hence, segmentation models like SAM do not facilitate open-vocabulary out of the box and need to be adapted to support the processing of textual information (e.g., by using additional models that generate text embeddings or models that provide bounding boxes as input such as Grounding DINO~\cite{GroundingDINO2023}). 

Early approaches in OVSS have been built upon standard zero-shot semantic segmentation, such as ZS3Net~\cite{ZS3Bucher2019}, using simple word2vec text encoders. 
Subsequent two-stage approaches made use of mask proposals based on MaskFormer~\cite{MaskFormer2021} in stage~1 followed by predictions of each mask by CLIP~\cite{ZSSeg2021,OVSeg2022,ZegFormer2022,MaskCLIPDing2022}. 
Recently, one-stage frameworks like SAN~\cite{SAN2023} generate masks in a side adapter network during the CLIP inference.
Therefore, CLIP does not classify many mask proposals but only the image ones, resulting in a faster inference. Other mask-based models like GroupViT~\cite{GroupViT2022} and ViewCo~\cite{ViewCo2023} are grouping pixels into larger segments which are then classified.

Decoder-focused approaches such as DenseCLIP~\cite{DenseCLIP2022} and LSeg~\cite{LSeg2022} encode the image with CLIP and obtain the pixel-level patch embeddings. Because the pre-training is focused on the class embedding, the approaches append additional decoders to refine the patch embeddings. For this refinement, CAT-Seg~\cite{CAT-Seg2023} utilizes multiple stages of cost aggregation to generate the final segmentation mask. PACL~\cite{PACL2022} aligns patch embeddings and class embeddings during training and, as a result, does not require segmentation-specific training data or additional modules.
Zero-shot semantic segmentation models have been combined with other tasks as well. OpenSeeD~\cite{OpenSeeD2023} implements open-vocabulary for object detection and segmentation. SEEM~\cite{SEEM2023} processes text prompts and additional inputs like visual prompts similar to SAM.
Apart from differences in the architecture, the models vary in the training process---particularly in fine-tuning CLIP's vision encoder. 

\subsection{Evaluation and benchmarking of zero-shot semantic segmentation} 
Zero-shot semantic segmentation models are typically evaluated on datasets consisting exclusively of everyday images, such as ADE20K~\cite{DatasetADE20K}, Pascal Context \citep{DatasetPascalContext}, and Pascal VOC~\cite{DatasetPascalVOC}. 
These dataset are the \textit{de facto} standard for evaluating these models (see~\cite{OVSeg2022, CAT-Seg2023, XDecoder2022, ZSSeg2021, OpenSeg2021, SAN2023}).
Few studies have considered additional datasets. Notably, \citet{XDecoder2022} proposed a Segmentation in the Wild (SegInW) benchmark with 25 datasets. However, the majority of the datasets in SegInW still consist of everyday images with only two exceptions: brain tumor segmentation and a bird's eye view in stables. 
To the best of our knowledge, zero-shot semantic segmentation and OVSS have not been evaluated on other datasets. 
Outside of zero-shot semantic segmentation and OVSS, semantic segmentation is usually evaluated based on collections of datasets, like MSeg~\cite{DatasetMSeg}. These datasets generally only include everyday images, indoor scenes, and driving datasets and lack domain-specific datasets. SAM has been evaluated on 23 instance segmentation datasets in a point-to-mask setting~\cite{SegmentAnything}. This collection of datasets is the most extensive for segmentation tasks but still misses domains, such as engineering and earth monitoring. Other works evaluate specifically domain dataset collections such as medical tasks~\cite{BenchmarkSAmedical} or satellite data~\cite{BenchmarkEarthMonitoring}. 

\section{MESS benchmark}

\begin{table}[b]
\caption{Multi-domain benchmark for zero-shot semantic segmentation models consisting of 22 downstream tasks, a total of 448 classes, and 25,079 images.
}\label{tab:taxonomy}

\begin{center}
\tiny
\setlength\extrarowheight{0.2em}
\addtolength{\tabcolsep}{-0.3em}

\begin{tabularx}{\textwidth}{l|ccYYYc|YY}
\toprule
Dataset & Domain & Sensor type & Segment size & Number of classes & Class similarity & Vocabulary & Number of images & Task\\ 
\midrule
BDD100K \cite{DatasetBDD100K} & \multirow{6}{*}{General} & Visible spectrum & Medium & 19 (Medium) & Low & Generic & 1,000 & Driving
\\
Dark Zurich \cite{DatasetDarkZurich} &  & Visible spectrum & Medium & 20 (Medium) & Low & Generic & 50 & Driving \\
MHP v1 \cite{DatasetMHP} &  & Visible spectrum & Small & 19 (Medium) & High & Task-spec. & 980 & Body parts
\\
FoodSeg103 \cite{DatasetFoodSeg103} &  & Visible spectrum &  Medium & 104 (Many) & High & Generic & 2,135 & Ingredients
\\
ATLANTIS \cite{DatasetATLANTIS} &  & Visible spectrum &  Small & 56 (Many) & Low & Generic & 1,295 &  Maritime
\\
DRAM \cite{DatasetDRAM} &  & Visible spectrum & Medium & 12 (Medium) & Low & Generic & 718 & Paintings \\
\hline
iSAID \cite{DatasetiSAID} & \multirow{5}{*}{\shortstack{Earth\\Monitoring}} & Visible spectrum & Small & 16 (Medium) & Low & Generic & 4,055 & Objects \\
ISPRS Potsdam \cite{DatasetISPRSPotsdam} &  & Multispectral & Small & 6 (Few) & Low & Generic & 504 & Land use \\
WorldFloods \cite{DatasetWorldFloods} &  & Multispectral & Medium & 3 (Binary) & Low & Generic & 160 & Floods\\
FloodNet \cite{DatasetFloodNet} &  & Visible spectrum & Medium & 10 (Few) & Low & Task-spec. & 5,571 & Floods \\
UAVid \cite{DatasetUAVid} &  & Visible spectrum & Small & 8 (Few) & High & Task-spec. & 840 & Objects\\
\hline
Kvasir-Inst. \cite{DatasetKvasirInstrument} & \multirow{4}{*}{\shortstack{Medical\\Sciences}} & Visible spectrum & Medium & 2 (Binary) & Low & Generic & 118 & Endoscopy\\
CHASE DB1 \cite{DatasetCHASEDB1} &  & Microscopic & Small & 2 (Binary) & Low & Domain-spec. & 20 & Retina scan \\
CryoNuSeg \cite{DatasetCryoNuSeg} &  & Microscopic & Small & 2 (Binary) & Low & Domain-spec. & 30 & WSI \\
PAXRay-4 \cite{DatasetPAXRay} &  & Electromagnetic & Large & 4x2 (Binary) & Low & Domain-spec. & 180 & X-Ray \\
\hline
Corrosion CS \cite{DatasetCorrosionCS} & \multirow{4}{*}{Engineering} & Visible spectrum & Medium & 4 (Few) & High & Task-spec. & 44 & Corrosion \\
DeepCrack \cite{DatasetDeepCrack} &  & Visible spectrum & Small & 2 (Binary) & Low & Generic & 237 & Cracks \\
ZeroWaste-f \cite{DatasetZeroWaste} &  & Visible spectrum & Medium & 5 (Few) & High & Generic & 929 & Conveyor \\
PST900 \cite{DatasetPST900} &  & Electromagnetic & Small & 5 (Few) & Low & Generic & 288 & Thermal \\
\hline
SUIM \cite{DatasetSUIM} & \multirow{3}{*}{\shortstack{Agriculture\\and Biology}} & Visible spectrum & Medium & 8 (Few) & Low & Generic & 110 & Underwater \\
CUB-200 \cite{DatasetCUB200} &  & Visible spectrum & Medium & 201 (Many) & High & Domain-spec. & 5,794 & Bird species \\
CWFID \cite{DatasetCWFID} &  & Visible spectrum & Small & 3 (Few) & High & Generic & 21 & Crops \\
\bottomrule
\end{tabularx}

\end{center}
\end{table}

Following the HELM benchmark~\cite{BenchmarkHELM} proposed for the evaluation of large language models, we develop a taxonomy with task characteristics for semantic segmentation and retrieve a set of more than 500 datasets that we review as part of the benchmark creation. 
For the development of the taxonomy, we use a method proposed by \citet{TaxonomyNickerson2013}. 
We start the development of the taxonomy by specifying the so-called meta-characteristic of the taxonomy (i.e., our goal): \textit{identify visual and language characteristics of downstream tasks influencing the performance of zero-shot semantic segmentation models}. 
We then initialize the taxonomy in a conceptual-to-empirical cycle based on a review of other benchmarks and literature. Next, we refine the taxonomy in multiple empirical-to-conceptual iterations. 
We reviewed semantic segmentation datasets on Papers with Code, Kaggle, and additional test datasets used by recent segmentation models.
We repeatedly reduced the dimensions of the taxonomy to the most meaningful ones for the meta-characteristic.
We then conducted a statistical analysis of potential taxonomy dimensions to identify and remove similar or overlapping dimensions (see supplementary material).
We identified multiple dimensions that highly correlate with each other like color map and sensor type, segment size and segments per image, as well as viewpoint and domain. Based on this analysis, we discarded color map, resolution, segments per image, and viewpoint.
The final taxonomy matches all ending conditions~\cite{TaxonomyNickerson2013}.
While the proposed taxonomy identifies the most important dimensions and characteristics validated based on 120 classified datasets, there may be additional dimensions that influence the performance of zero-shot semantic segmentation models in specific cases.  Overall, we observe that certain characteristics are more likely to co-occur. E.g, binary datasets typically imply a low class similarity, whereas task-specific vocabulary is often associated with a high similarity between the task-specific classes. We account for this imbalance in the distribution of the characteristics and reflect it in our benchmark.

Following the taxonomy development, we selected a representative set of datasets so that the MESS benchmark is informative, reproducible, and manageable. Specifically, we filtered the 120 classified datasets based on four exclusion criteria: each dataset has an official and annotated validation or test set, high annotation quality, moderate disk usage, and sufficient image size. Next, we selected a subset that consists of complementing use cases to avoid duplication and covers all characteristics of the taxonomy. We present the 22 selected datasets and their characteristics within the taxonomy’s dimensions in Table~\ref{tab:taxonomy} . 
These datasets cover a variety of applications, resulting in a holistic evaluation of domain-specific applications. We publish this new MESS benchmark at \url{https://blumenstiel.github.io/mess-benchmark} and invite others to suggest additional datasets and refine classes for future versions.

During dataset selection, we have not identified any ethical issues with these datasets based on the information provided by the data sources. Our use follows the terms and conditions set by the data providers, and we list the corresponding licenses in the supplementary material. However, we acknowledge the importance of considering the societal impact of our work. FMs, such as CLIP, are pre-trained on vast corpora of data that may contain biases. We refer to \citet{EvaluatingCLIP2021} for a detailed analysis of biases in CLIP. While the majority of MESS datasets are less prone to such biases, some may include data specific to gender or geographic regions. We believe that assessing model performance across a range of datasets can help to identify and mitigate the impact of biases.

\section{Experimental setup}

In this section, we provide a brief definition of the zero-shot semantic segmentation task, describe the metrics, and outline implementation details.

\subsection{Task}

Let $I$ denote an image with a set of candidate classes $\mathcal{C} = \{C_1, C_2, ..., C_N\}$, where each candidate class $C_i$ is described in natural language. Zero-shot semantic segmentation models then assign a class $C_i$ to each pixel of $I$. The number of candidate classes $N$ can vary during inference (e.g., different downstream tasks) and, additionally, the model may not have seen the candidate classes during training. This is in contrast to traditional semantic segmentation, where the set of classes is fixed during training and inference~\cite{CAT-Seg2023}.
Each dataset represents a set of images with the same label set, and 
in our evaluations, none of the models is trained on the datasets from the benchmark or the same set of candidate classes. 
However, it is reasonable to assume the evaluated classes have been present in the pre-training of the underlying vision-language models (like CLIP).
All evaluated models have been trained on images with three channels (i.e., RGB). To account for datasets with varying numbers of input channels we mapped them to RGB (i.e., inputs with a single channel are mapped to RGB, for multispectral inputs we selected a subset of three channels).

\subsection{Implementation}
Following common practice, we evaluate all models using the mean of class-wise intersection over union (mIoU)~\cite{CAT-Seg2023, SAN2023, OVSeg2022, ZSSeg2021, MaskFormer2021}. 
%
We split very large images from the earth monitoring datasets into smaller patches of 1024$\times$1024 pixels. 
Further, we use an IRRG color map for multispectral datasets (ISPRS Potsdam and WorldFloods) and select the thermal data in PST900. All other datasets include images with one or three channels.
Across our implementation, we use PyTorch~\cite{Pytorch2019} and Detectron2~\cite{Detectron22019} for implementing the data loaders. For the convenience of users and contributors to our benchmark, we additionally provide wrappers for torchvision and MMSeg to process datasets in the Detectron2 dataset catalog. 
We did not train any models but used the publicly available weights and model configurations. The evaluation was conducted on an NVIDIA V100S.

\subsection{Models}\label{subsec:models}
We utilize our MESS benchmark to evaluate a range of recent models for zero-shot semantic segmentation including the state-of-the-art,
selecting models based on the reported performance and the availability of official code and weights.
OVSeg~\cite{OVSeg2022}, SAN~\cite{SAN2023}, and CAT-Seg~\cite{CAT-Seg2023} represent the state-of-the-art across different approaches in the architecture for zero-shot semantic segmentation (i.e., two-stage mask-based, one-stage mask-based, and pixel-based). 
We additionally consider ZSSeg~\cite{ZSSeg2021} and ZegFormer~\cite{ZegFormer2022} which are frequently consulted as baseline models e.g. by~\cite{OVSeg2022, CAT-Seg2023, SAN2023}.
The previously listed models are based on CLIP and use COCO Stuff to train the additional segmentation modules. Additionally, OVSeg uses COCO Captions for fine-tuning. 
X-Decoder~\cite{XDecoder2022} and OpenSeeD~\cite{OpenSeeD2023} are part of our evaluation since these approaches do not make use of CLIP but are based on UniCL~\cite{UniCL2022} (i.e., their public versions). X-Decoder and OpenSeeD are trained on multiple datasets which we detail in the supplementary material.

To account for recent developments in the field, we additionally include SAM~\cite{SegmentAnything} in our evaluations. Standard SAM can only process visual prompts and does not facilitate text-to-mask settings.
Therefore, we validated other ways to make use of SAM. We implement Grounded-SAM~\cite{GroundedSAM2023} using the predicted bounding boxes from Grounding DINO~\cite{GroundingDINO2023} as input for SAM and thereby enabling an open-vocabulary setting (i.e., text-to-mask). This serves as a baseline to better understand the potential of SAM-based text-to-mask models. 
The overall evaluation time per model on the MESS benchmark varies in our experiments between 1 hour for SAN-B and 14.5 hours for OVSeg-L.

\section{Experiments}\label{sec:results}

In the following, we provide a holistic overview of the performance of multiple zero-shot semantic segmentation models based on our MESS benchmark.
We conduct a range of in-detail analyses of model performances across the dimensions of our taxonomy including sensor types, the class similarity, and the vocabulary---additional experiments are included in the supplementary material. 

\begin{table}[tbh]
\caption{mIoU results averaged by the dataset domain. Best-performing models are highlighted in bold, and the second-best are underlined. Random represents the randomly expected mIoU with uniformly distributed predictions. The best supervised models are separately selected for each dataset (see supplementary material for the supervised models and results).}\label{tab:results}

\begin{center}
\setlength\extrarowheight{0.1em}
\addtolength{\tabcolsep}{-0.4em}
\small
\begin{tabularx}{\linewidth}{lYY|YYYYY|Y}
\toprule
Model & Parameters & Inference (s/iter) & General & Earth Monit. & Medical Sciences & Engineer. & Agri. and Biology & Mean \\
\midrule
\textit{Random (LB)} & & & \phantom{0}\textit{1.17} & \phantom{0}\textit{7.11} & \textit{29.51} & \textit{11.71} & \phantom{0}\textit{6.14} & \textit{10.27} \\
\multicolumn{2}{l}{\textit{Best supervised (UB)}} & & \textit{48.62} & \textit{79.12} & \textit{89.49} & \textit{67.66} & \textit{81.94} & \textit{70.99} \\
\hline
\hline
ZSSeg-B~\cite{ZSSeg2021} & 211M & 0.49 & 19.98 & 17.98 & \underline{41.82} & 14.0\phantom{0} & 22.32 & 22.73 \\
ZegFormer-B~\cite{ZegFormer2022} & 210M & 0.18 & 13.57 & 17.25 & 17.47 & 17.92 & \underline{25.78} & 17.57 \\
X-Decoder-T~\cite{XDecoder2022} & 164M & 0.1 & 22.01 & 18.92 & 23.28 & 15.31 & 18.17 & 19.8\phantom{0} \\
SAN-B~\cite{SAN2023} & 158M & 0.04 & \underline{29.35} & \underline{30.64} & 29.85 & \underline{23.58} & 15.07 & \underline{26.74} \\
OpenSeeD-T~\cite{OpenSeeD2023} & 116M & 0.08 & 22.49 & 25.11 & \textbf{44.44} & 16.5\phantom{0} & 10.35 & 24.33 \\
CAT-Seg-B~\cite{CAT-Seg2023} & 181M & 0.17 & \textbf{34.96} & \textbf{34.57} & 41.65 & \textbf{26.26} & \textbf{29.32} & \textbf{33.74} \\
\hline
OVSeg-L~\cite{OVSeg2022} & 531M & 1.64 & 29.54 & 29.04 & 31.9\phantom{0} & 14.16 & 28.64 & 26.94 \\
SAN-L~\cite{SAN2023} & 437M & 0.14 & 36.18 & \underline{38.83} & 30.27 & 16.95 & 20.41 & 30.06 \\
CAT-Seg-L~\cite{CAT-Seg2023} & 490M & 0.33 & \textbf{39.93} & \textbf{39.85} & \textbf{48.49} & \underline{26.04} & \underline{34.06} & \textbf{38.14} \\
CAT-Seg-H~\cite{CAT-Seg2023} & 1049M & 0.5 & \underline{37.98} & 37.74 & \underline{34.65} & \textbf{29.04} & \textbf{37.76} & \underline{35.66} \\
\bottomrule
\end{tabularx}

\label{tab:domain_results_mIoU}
\end{center}
\end{table}

\begin{figure}[bh]
    \centering    
    \begin{minipage}[t]{.48\textwidth}    
    \includegraphics[width=\linewidth]
    {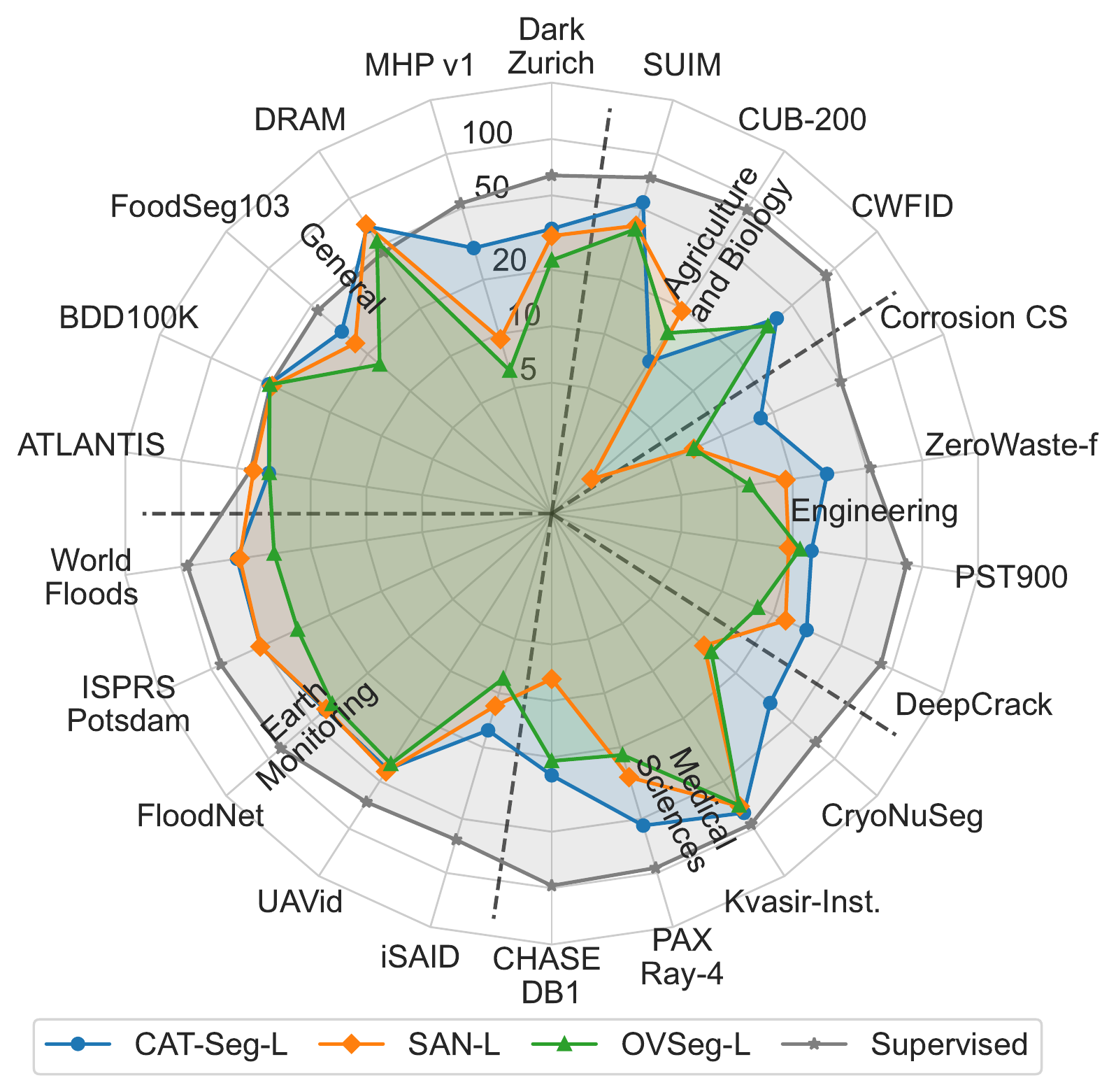}
    \caption{mIoU results for large models on a log scale. The datasets are grouped by their domain and sorted by supervised performance.}\label{fig:radar}
    \end{minipage}
    \hspace*{\fill} 
    \begin{minipage}[t]{.48\textwidth}    
    \includegraphics[width=\linewidth]{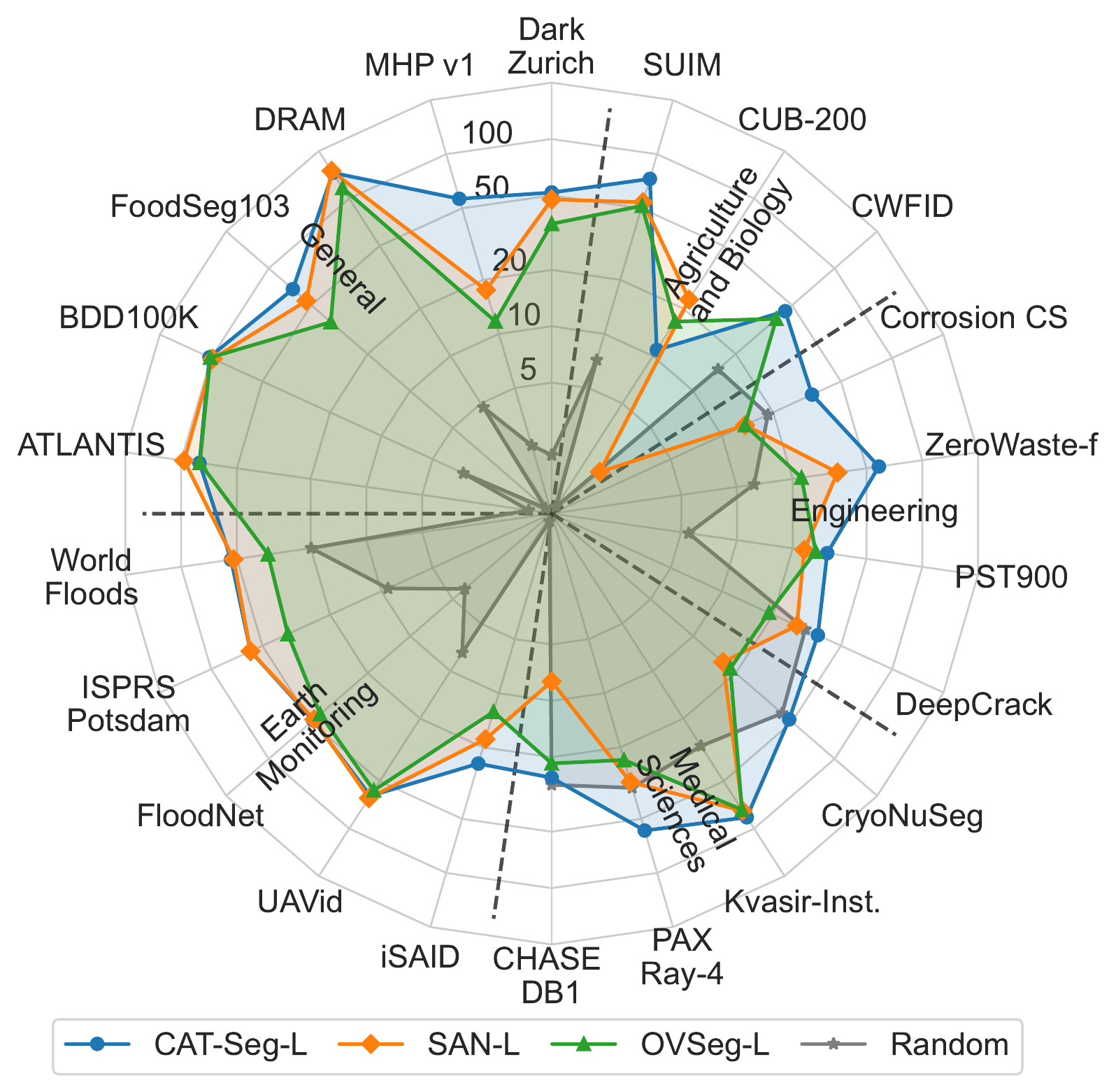}
    \caption{mIoU results of large models in relative comparison to the supervised mIoU on a log scale. 100 is equal to the supervised mIoU.}\label{fig:radar_supervised_delta}
    \end{minipage}
\end{figure}

\subsection{Multi-domain zero-shot semantic segmentation}

We provide a quantitative comparison across models and all datasets summarized by their domain in Table \ref{tab:results} and per dataset results in Fig.~\ref{fig:radar} and \ref{fig:radar_supervised_delta}. 
We add a random prediction as a lower bound by calculating the expected mIoU value with uniformly distributed predictions over all classes. 
In addition, we report fully supervised results based on the current SOTA from supervised semantic segmentation (see supplementary material).
Overall, CAT-Seg-L achieves a strong performance across domains with an average mIoU of 38.14\%, followed by its base and huge version. CAT-Seg is followed by SAN-L with a performance of 30.06\%. Notably, the performance of zero-shot CAT-Seg-L in the general domain is only 8.69pp (average mIoU) below the performance of supervised SOTA approaches. In comparison, CAT-Seg-L reaches on average 50.36\% of the supervised performance in earth monitoring and 54.18\% on medical sciences. The performance gap compared to supervised models is even larger for the two other domains. Looking at the dataset-specific performance in Fig.~\ref{fig:radar} and \ref{fig:radar_supervised_delta}, we observe that the performance varies between datasets and models. While SAN-L is the best-performing model on CUB-200 and DRAM, it has significantly lower performance on CWFID or CHASE DB1 compared to CAT-Seg-L. The model achieves scores between 50\% and over 100\% of the performance of supervised state-of-the-art in the general domain. Within the other domains, CAT-Seg-L has a performance gap of more than 25pp for most of the datasets.

The inference time varies between the models and, in particular, between different model architectures with some models requiring more than ten times higher computational effort indicated by higher inference times.  In general, we observe the highest inference times for two-stage mask-based approaches, such as ZSSeg and OVSeg, which are between five to twelve times higher than other mask-based approaches (X-Decoder, OpenSeeD, and SAN). The point-based CAT-Seg uses a sliding window approach which requires five passes and therefore results in higher inference times than SAN. Overall, SAN represents the fastest model in our experiments.

\begin{table}[b]
\caption{Comparision of mIoU results for images with different sensor types. Pseudo refers to thermal data mapped to a pseudo color map.}\label{tab:sensor_type}

\begin{center}
\setlength\extrarowheight{0.1em}
\addtolength{\tabcolsep}{-0.4em}
\small
\begin{tabularx}{\linewidth}{lYYYYYYY}
\toprule
 & \multicolumn{2}{c}{ISPRS Potsdam} & \multicolumn{2}{c}{WorldFloods} & \multicolumn{3}{c}{PST900} \\
Model & IRRG & RGB & IRRG & RGB & Thermal & Pseudo & RGB \\
\midrule
OVSeg-L~\cite{OVSeg2022} & 31.03 & 35.46 & 31.48 & 22.86 & \underline{21.89} & \underline{21.63} & 42.9\phantom{0} \\
SAN-L~\cite{SAN2023} & \textbf{51.45} & \textbf{52.06} & \underline{48.24} & \textbf{45.93} & 19.01 & 19.41 & \underline{49.02} \\
CAT-Seg-L~\cite{CAT-Seg2023} & \underline{51.42} & \underline{51.29} & \textbf{49.86} & \underline{45.39} & \textbf{25.26} & \textbf{25.43} & \textbf{65.55} \\
\bottomrule
\end{tabularx}

\end{center}
\end{table}

\subsection{Sensor type evaluation}

All considered models have been developed for the visual spectrum (i.e., RGB).  In the following, we investigate the performance of three different sensor types: multispectral, electromagnetic, and microscopic.
Three datasets from MESS allow for a direct comparison between different sensor types. For multispectral sensors, the MESS benchmark includes the IRRG color map for ISPRS Potsdam and WorldFloods. 
The models are able to process the different color maps and profit from the visual highlighting of vegetation through the infrared channel. This insight might be limited to commonly used color maps because other color maps might be less represented in the pre-training data of CLIP.
On electromagnetic and thermal imagery, none of the evaluated models can regularly segment objects on the PST900 dataset. 
We compared this result to the aligned RGB images from PST900. 
All models perform significantly better on the RGB images. E.g., CAT-Seg-L reaches a mIoU of 65.55\% on RGB images compared to only 25.26\% for thermal data. We also tested a pseudo color map that maps the grayscale thermal data to a pseudo color scale, resulting in a similar low performance. Therefore, we conclude that zero-shot semantic segmentation models are currently not able to sufficiently segment objects in thermal images. Most models are also not able to correctly segment X-ray images in the PAXRay dataset, the second benchmark dataset with an electromagnetic sensor type. However, X-rays do include much more visual features compared to thermal images and CAT-Seg is able to segment some anatomical structures like the lungs.
Further, the benchmark includes retina scans in CHASE DB1 and WSI images in CrypNuSeg to evaluate microscopic imagery. Similar to the PAXRay results, most models fail to segment the structures. But CAT-Seg and ZSSeg can locate the requested class. Thus, we assume that CLIP and zero-shot semantic segmentation can understand microscopic concepts but the correct segmentation is not achieved because of the small segments instead of the image type.

\subsection{Multi-domain vs. in-domain evaluation}
\begin{figure}[tbh]
    \centering    
    \begin{minipage}[t]{.48\textwidth} 
    \centering
    \includegraphics[width=0.8\textwidth]{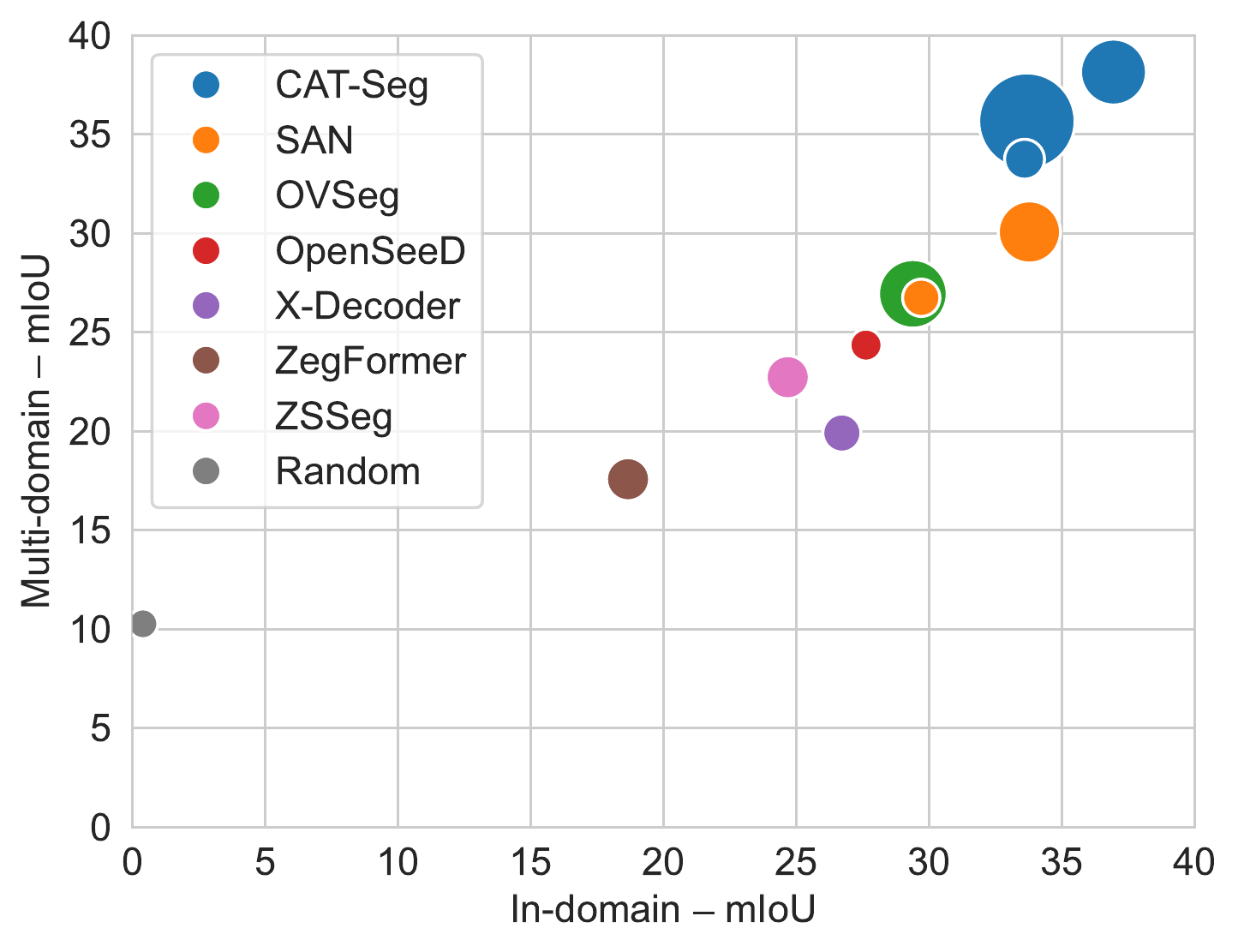}
  \caption{Cross domain settings can be challenging: Average mIoU of commonly used evaluation datasets in comparison to the results on the MESS benchmark. The size represents the parameter count of the models.}\label{fig:common_domain}
    \end{minipage}
    \hspace*{\fill} 
    \begin{minipage}[t]{.48\textwidth}  
    \centering
     \includegraphics[width=0.8\textwidth]{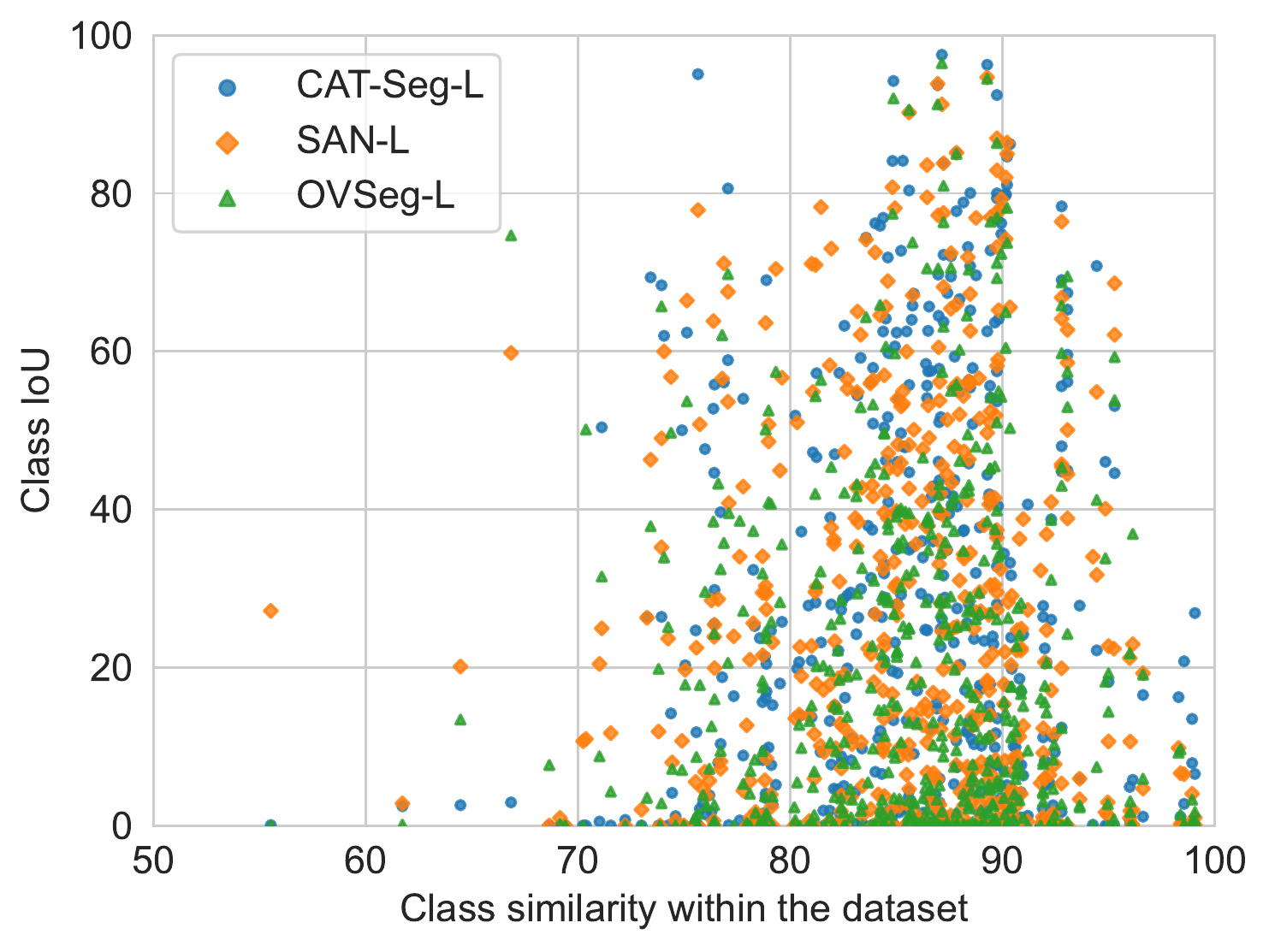}
  \caption{The class-wise IoU in comparison with the similarity to other labels within the dataset. The similarity is measured by the minimum cosine distance of the class label to all other CLIP text embeddings within the dataset.}\label{fig:iou_similarity}
    \end{minipage}
\end{figure}

Most zero-shot semantic segmentation models are currently evaluated on five datasets: Pascal VOC, ADE20K-150, ADE20K-847, Pascal Context-59, and Pascal Context-459. Figure \ref{fig:common_domain} compares the average results of the evaluated models on these common datasets (i.e., in-domain datasets) to a multi-domain setting with datasets of MESS benchmark. Note that the multi-domain datasets contain fewer classes on average, resulting in a much higher random mIoU. We provide the results for each dataset in the supplementary material. While SAN-L has comparable performance to the CAT-Seg models on common datasets, it has a significantly lower mIoU on domain datasets. Further, X-Decoder has a generally lower mIoU on domain datasets compared to other models. X-Decoder does not use CLIP which may explain the limited generalizability of the model. 
Overall, CAT-Seg is the only model architecture with a higher average mIoU on the domain datasets than common datasets.

\subsection{Language characteristics}
The differentiation between related classes is relevant in domain-specific use cases like biology. We analyze the influence of class similarity on class-wise IoU in Figure \ref{fig:iou_similarity}. Following~\citet{SAN2023}, we calculated the class similarity as the maximum cosine similarity of the embedding to all other CLIP text embeddings in the label set.
Overall, the class IoU does not correlate with the similarity. However, none of the classes with high similarity reaches a desirable IoU (e.g., the Corrosion CS dataset with three classes describing different corrosion stages). 
All models face difficulties in differentiating these classes.  
In additional experiments, the model performance significantly improved when considering similar classes as a single class.
Also, specialized terms affect the model performance, specifically, domain-specific and task-specific labels.
Our evaluation covers domain-specific words from medicine and biology, i.e., bird species and anatomical structures like the \textit{mediastinum}. It shows that CLIP is able to understand domain-specific concepts to a limited extent. We observed higher performance for generic terminology. 
E.g., all models achieve higher performances on the Kvasir-Instrument dataset when using a generic vocabulary like \textit{tool}. Utilizing a more precise term like \textit{surgical instrument} reduces the mIoU. 
%
We refer to classes with specified conditions as task-specific classes. 
In our evaluations, CAT-Seg achieves the best results on task-specific classes. 
However, CAT-Seg still confuses classes and, e.g., predicts the right shoe and right leg significantly more often than the left side in MHP v1. 
CAT-Seg models are further biased towards the \textit{parked car} class in UAVid images, while SAN and OVSeg mostly assign masks to the label \textit{moving car}.
Overall, domain-specific and task-specific vocabulary limits the performance of zero-shot semantic segmentation models. 

\begin{table}[tbh]
\caption{Domain-averaged mIoU results for Grounded-SAM and SAM with oracle inputs in a point-to-mask and box-to-mask setting. Random, supervised and CAT-Seg-L are provided for reference.}\label{tab:results_sam}

\begin{center}
\setlength\extrarowheight{0.1em}
\addtolength{\tabcolsep}{-0.4em}
\small
\begin{tabularx}{\linewidth}{lY|YYYYY|Y}
\toprule
Model & Input prompt & General & Earth Monitoring & Medical Sciences & Engineering & Agri. and Biology & Mean \\
\midrule
\textit{Random (LB)} & & \phantom{0}\textit{1.17} & \phantom{0}\textit{7.11} & \textit{29.51} & \textit{11.71} & \phantom{0}\textit{6.14} & \textit{10.27} \\
\multicolumn{2}{l|}{\textit{Best supervised (UB)}} & \textit{48.62} & \textit{79.12} & \textit{89.49} & \textit{67.66} & \textit{81.94} & \textit{70.99} \\
CAT-Seg-L~\cite{CAT-Seg2023} & & 39.93 & 39.85 & 48.49 & 26.04 & 34.06 & 38.14 \\
\hline
\hline
Gr.-SAM-B~\cite{GroundedSAM2023} & \multirow{3}{*}{\shortstack{Grounding\\DINO~\cite{GroundingDINO2023}}} & 29.51 & 25.97 & 37.38 & \textbf{29.51} & 17.66 & 28.52 \\
Gr.-SAM-L~\cite{GroundedSAM2023} & & \textbf{30.32} & \textbf{26.44} & \textbf{38.69} & \underline{29.25} & \textbf{17.73} & \textbf{29.05} \\
Gr.-SAM-H~\cite{GroundedSAM2023} & & \underline{30.27} & \textbf{26.44} & \underline{38.45} & 28.16 & \underline{17.67} & \underline{28.78} \\
\hline
SAM-B~\cite{SegmentAnything} & \multirow{3}{*}{\shortstack{Oracle\\points~\cite{RITM2022}}} & \textbf{50.41} & \underline{38.72} & 43.7\phantom{0} & 45.16 & \underline{57.84} & \underline{46.59} \\
SAM-L~\cite{SegmentAnything} & & \underline{45.99} & \textbf{44.03} & \underline{55.74} & \textbf{50.0}\phantom{0} & \textbf{58.23} & \textbf{49.99} \\
SAM-H~\cite{SegmentAnything} & & 36.05 & 34.82 & \textbf{59.58} & \underline{47.35} & 39.91 & 43.0\phantom{0} \\
\hline
SAM-B~\cite{SegmentAnything} & \multirow{3}{*}{\shortstack{Oracle\\bounding\\boxes}} & \textbf{78.5}\phantom{0} & \textbf{73.56} & \textbf{68.14} & \textbf{73.29} & \underline{86.0}\phantom{0} & \textbf{75.67} \\
SAM-L~\cite{SegmentAnything} &  & \underline{78.0}\phantom{0} & \underline{73.27} & 64.98 & \underline{73.09} & \textbf{86.99} & \underline{74.97} \\
SAM-H~\cite{SegmentAnything} &  & 65.23 & 59.61 & \underline{66.58} & 66.4 & 78.63 & 66.55 \\
\bottomrule
\end{tabularx}

\end{center}
\label{tab:domain_results_sam_mIoU}
\end{table}

\subsection{Comparison to SAM}

For a better understanding of current text-to-mask zero-shot semantic segmentation approaches, we compare them with grounded and oracle versions of SAM. SAM cannot directly process textual inputs, instead, it uses visual prompt inputs, i.e., bounding boxes or points. 
For the comparison, we implemented three versions of SAM. First, we made use of existing available demos combining Grounding DINO and SAM and extended them by a comprehensive quantitative evaluation. 
Second, oracle point-to-mask SAM refers to a model that provides a single point for every connected segment in the ground truth mask to simulate the visual input. We use the point sampling approach from RITM~\cite{RITM2022}.
Third, oracle box-to-mask SAM utilizes a single box for every segment in the ground truth mask to simulate the visual input. We consider up to 100 input prompts per image to avoid a large number of very small segments.
We later combine all predicted masks by taking the maximum logit value for each pixel. Pixels with only negative logit values are assigned to the background class or marked \textit{unlabeled} in datasets without a background class.
Note that inputting text data as in the models before is fundamentally different from utilizing visual inputs as in our two oracle SAM implementations and our analyses are not intended for a direct comparison but to better understand the potentials of SAM for zero-shot text-to-mask models.

In Table~\ref{tab:domain_results_sam_mIoU}, we observe that the non-oracle implementation of SAM utilizing Grounding DINO generally exhibits limited performance compared to CAT-Seg text-to-mask models. Oracle versions of SAM receive significantly improved information on the location of the object and, therefore, show a strong performance. Given the perfect information on the location of objects in the image with oracle bounding boxes, the oracle box-to-mask SAM implementation even outperforms supervised semantic segmentation models. Overall, we observe that SAM models achieve a strong performance based on oracle information on the location of the objects. However text-to-mask zero-shot semantic segmentation models like CAT-Seg outperform the combination of Grounding DINO and SAM. Similar to X-Decoder and OpenSeeD, Grounding DINO does not use CLIP, which results in limited multi-domain performance. The results with oracle bounding boxes suggest that future combinations of SAM with open-vocabulary object detection models based on FMs like CLIP may outperform the current state-of-the-art in zero-shot semantic segmentation.

\subsection{Qualitative analyses}

In the following, we quantitatively compare the predictions of the three promising text-to-mask zero-shot semantic segmentation models with the ground truth and the grounding version of SAM on four different datasets (autonomous driving, satellite imagery, medical science, and engineering). We visually observe the following characteristics: First, CAT-Seg also visually surpasses the predictions of the other models. Second, across different domains, the predictions of CAT-Seg are largely in line with the ground truth and the segmentation is comparatively fine-grained. Third, we observe that Grounding DINO does not locate most segments and, therefore, Grounded-SAM tends to predict the background class. These qualitative observations are largely in line with our quantitative experiments.

\begin{figure}[tbh]
    \centering
    \footnotesize
    \addtolength{\tabcolsep}{-0.5em}    \begin{tabular}{lcccccc}
 & Image & Ground Truth & CAT-Seg-L & SAN-L & OVSeg-L & Gr.-SAM-L \\
\rotatebox[origin=l]{90}{Dark Z.} & \includegraphics[width=.15\textwidth]{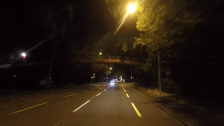} & \includegraphics[width=.15\textwidth]{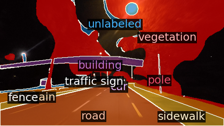} & \includegraphics[width=.15\textwidth]{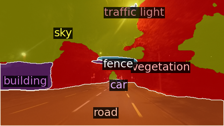} & \includegraphics[width=.15\textwidth]{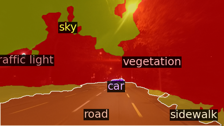} & \includegraphics[width=.15\textwidth]{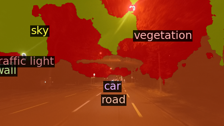} & \includegraphics[width=.15\textwidth]{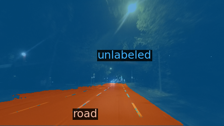} \\
\rotatebox[origin=l]{90}{WorldFloods} & \includegraphics[width=.15\textwidth]{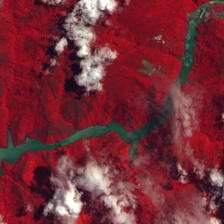} & \includegraphics[width=.15\textwidth]{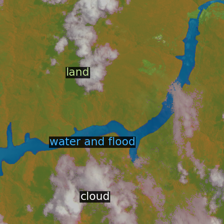} & \includegraphics[width=.15\textwidth]{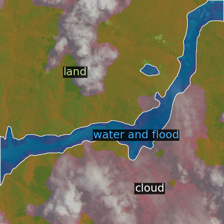} & \includegraphics[width=.15\textwidth]{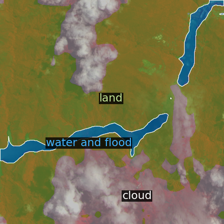} & \includegraphics[width=.15\textwidth]{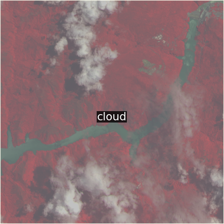} & \includegraphics[width=.15\textwidth]{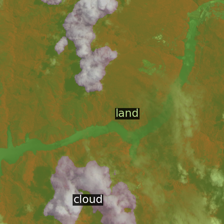} \\
\rotatebox[origin=l]{90}{CryoNuSeg} & \includegraphics[width=.15\textwidth]{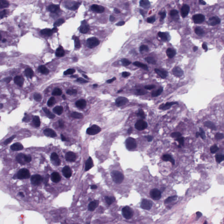} & \includegraphics[width=.15\textwidth]{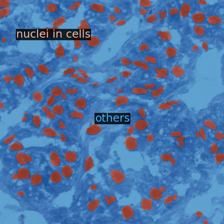} & \includegraphics[width=.15\textwidth]{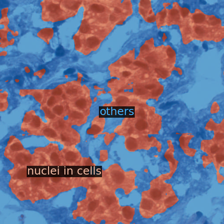} & \includegraphics[width=.15\textwidth]{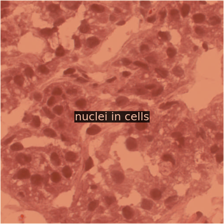} & \includegraphics[width=.15\textwidth]{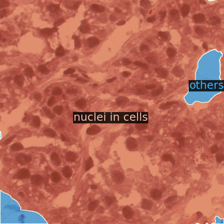} & \includegraphics[width=.15\textwidth]{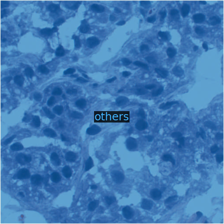} \\
\rotatebox[origin=l]{90}{Corr. CS} & \includegraphics[width=.15\textwidth]{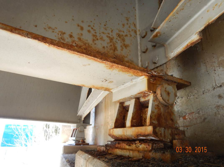} & \includegraphics[width=.15\textwidth]{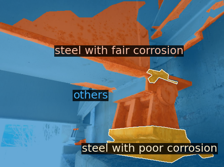} & \includegraphics[width=.15\textwidth]{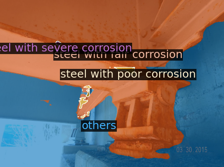} & \includegraphics[width=.15\textwidth]{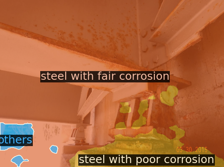} & \includegraphics[width=.15\textwidth]{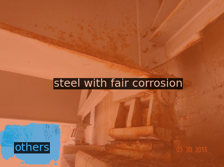} & \includegraphics[width=.15\textwidth]{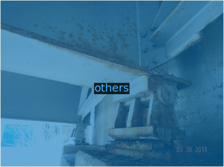} \\
\end{tabular}

    \caption{Predictions from selected datasets based on CAT-Seg-L~\cite{CAT-Seg2023}, SAN-L~\cite{SAN2023}, OVSeg-L~\cite{OVSeg2022}, and Grounded-SAM~\cite{GroundedSAM2023}.}
    \label{fig:predictions}
\end{figure}

Zero-shot semantic segmentation achieves a remarkable performance on in-domain datasets \cite{CAT-Seg2023, SAN2023}. Based on the MESS benchmark, we observe that these models can solve some tasks from other domains, however, are limited in their applicability to domains like medical science, engineering, and agriculture. 
We identified a range of challenges: First, we observe that domain-specific and task-specific vocabulary are difficult to handle.
Models tend to be confused by labels with a high class similarity as in Corrosion CS. 
Therefore, we recommend to utilize a generic vocabulary with common class names, which led to improved performances in our experiments (e.g., \textit{tool} instead of \textit{medical instrument} in Kvasir-Instrument). 
Second, differences in the type of the sensor influence the performance of these models which are generally trained on the visual spectrum---for example, thermal data is hard to process.
Third, we observe that state-of-the-art text-to-mask approaches outperform Grounded-SAM across multiple domains.

\section{Conclusion}

Zero-shot semantic segmentation has the potential to make segmentation models more accurate, cheap, flexible, and interactive. However, the current evaluation is limited to in-domain datasets, and previous analyses focused on model properties rather than task characteristics.
With the MESS benchmark, we enable a holistic evaluation and 
invite others to utilize this benchmark to accelerate the field of semantic segmentation across domains to improve its real-world applicability.

\section{Acknowledgements}

We want to acknowledge the prior work this benchmark builds on. We especially want to emphasize that we leverage works across the AI community that should be recognized and cited. We appreciate the significant effort across the community in the careful collection, annotation, and publication of datasets. In the code repository, we provide additional details of the datasets including links to the corresponding works for citation. Additionally, we want to be explicit that our evaluation across diverse approaches would not be possible without publicly available architectures and corresponding model weights.

\bibliographystyle{apalike} 
\bibliography{references.bib}

\end{document}


\maketitle

\appendix

The supplementary material is organized as follows: 
\begin{itemize}
    \item We detail the taxonomy development in Section \ref{sec:taxonomy}.
    \item The benchmark datasets are analyzed in Section \ref{sec:datasets}.
    \item We provide details about the evaluated models in Section \ref{sec:models}.
    \item Additional experiments are presented in Section \ref{sec:experiments}.
    \item Exemplary predictions are included in Section \ref{sec:predictions}.
\end{itemize}

\section{Taxonomy development}\label{sec:taxonomy}
The taxonomy and the characterized datasets serve as a basis for the selection of the benchmark datasets.
Therefore, we describe the taxonomy development in this section in detail. We applied the taxonomy development method proposed by \citet{TaxonomyNickerson2013} to analyze the task space of semantic segmentation. The method aims to develop a framework based on deduction and induction rather than \textit{ad-hoc} decisions. We initialize the development by selecting our meta-characteristic (i.e., the goal): \textit{identify visual and language characteristics of downstream tasks influencing the performance of zero-shot semantic segmentation models}.

We apply multiple empirical-to-conceptual or conceptual-to-empirical cycles until the ending conditions are reached. In an empirical-to-conceptual iteration, new objects are examined, and common characteristics are identified. The characteristics must derive from the meta-characteristic and discriminate between the objects to be of use for the taxonomy. 
In the conceptual step, the characteristics are grouped into dimensions. In contrast, a conceptual-to-empirical cycle starts by deducting potential dimensions and characteristics for the meta-characteristic based on prior knowledge. Next, the concept is evaluated by classifying objects. 
If a dimension does not differentiate between the objects or a characteristic has no real examples, it might not be appropriate. 
To fulfill the subjective ending conditions, the taxonomy must be concise, robust, comprehensive, extendible, and explanatory. Further, the objective ending conditions include, among others, dimension uniqueness and characteristic uniqueness within the dimension. We refer to \citet{TaxonomyNickerson2013} for more detailed information.

Starting with a first conceptual-to-empirical cycle, we analyzed other benchmarks and literature to initialize the taxonomy. The RF100 object detection benchmark \citep{BenchmarkRoboflow100} clusters the datasets into seven categories, representing different image types. We analyzed the literature of the RF100 categories to identify visual or language features relevant to these images. Aerial and satellite imagery has many important characteristics, such as the \textit{sensor type} with different bands that can be mapped to true or false \textit{color maps}, the \textit{spatial resolution}, and \textit{metadata} like time and location \citep{TaxonomyPradham2008}. Electromagnetic images are often medical modalities using different sensor types, different \textit{sensor directions} and having a domain-specific \textit{vocabulary} that describes the anatomy \citep{TaxonomyLehmann2003}. Various sensors are also used in underwater imagery, and \textit{preprocessing} plays an essential role in this image domain \citep{TaxonomyLu2017}. The spatial resolution is an important factor in microscopy, besides sensor types and specific hardware such as phase contrast or fluorescence \citep{TaxonomyLiu2021}. Images from documents or video games are often synthetic but use the same visual spectrum as real-world images. Segmentation in documents does include very fine \textit{segment sizes}. 
We further added a \textit{domain} dimension because the image and labels are often domain-specific. Therefore, we select domains inspired by the major subject areas from Scopus\footnote{List of subject areas: https://www.scopus.com/sources} but shorten the labels to improve the usability of the taxonomy. We added a domain "General" for everyday images, which we did not associate with any subject area. 
Based on this research, we used all identified dimensions relevant to at least two domains as domain-specific dimensions lead to redundancy in the taxonomy. The initial characteristics of each dimension are selected based on the literature review and complemented through the following empirical step.

We refine the taxonomy in multiple empirical-to-conceptual iterations. Therefore, we reviewed overall 500 datasets, including all semantic segmentation datasets on Papers with Code\footnote{Semantic segmentation datasets: https://paperswithcode.com/datasets?task=semantic-segmentation}, Kaggle\footnote{Seach results for "semantic segmentation": https://www.kaggle.com/datasets?search=semantic+segmentation}, and test datasets from other segmentation models like SAM~\cite{SegmentAnything}. 
We classified 120 datasets within our taxonomy which are presented in Table \ref{tab:classification}. Note, that different versions of a dataset are classified if they lead to varying characteristics. We did not classify all datasets of similar use cases as we aim for a diverse collection of datasets (e.g., seven driving datasets out of 30+). Other criteria for exclusion are deviating tasks (e.g., 3D data) and missing data availability. Also, we discarded use cases that seem to be very unique, like galaxy segmentation. 
If only a few reviewed datasets covered specific use cases, e.g., crack segmentation, we analyzed additional datasets from other sources. 
Based on the datasets, we added dimensions regarding the segmentation mask and language-related dimensions like the \textit{class similarity}. We repeatedly reduced the dimensions in conceptual phases to the most meaningful ones for the meta-characteristic.

\begin{figure}[bt]
    \centering
    \subfloat[\centering Interim dimensions]{{\includegraphics[width=0.45\textwidth]{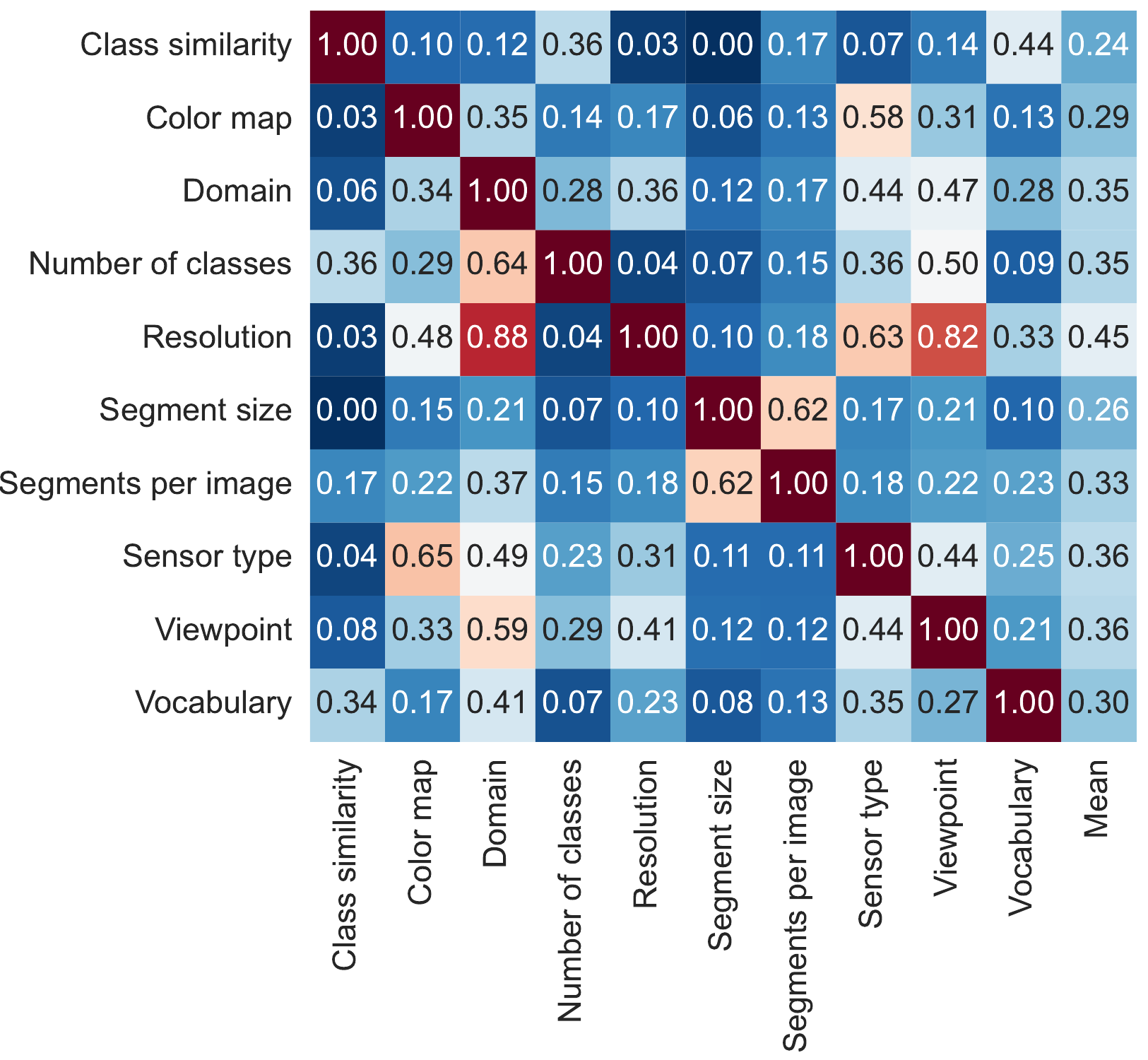} }}
    \hfill
    \subfloat[\centering Final dimension]
    {{\includegraphics[width=0.45\textwidth]{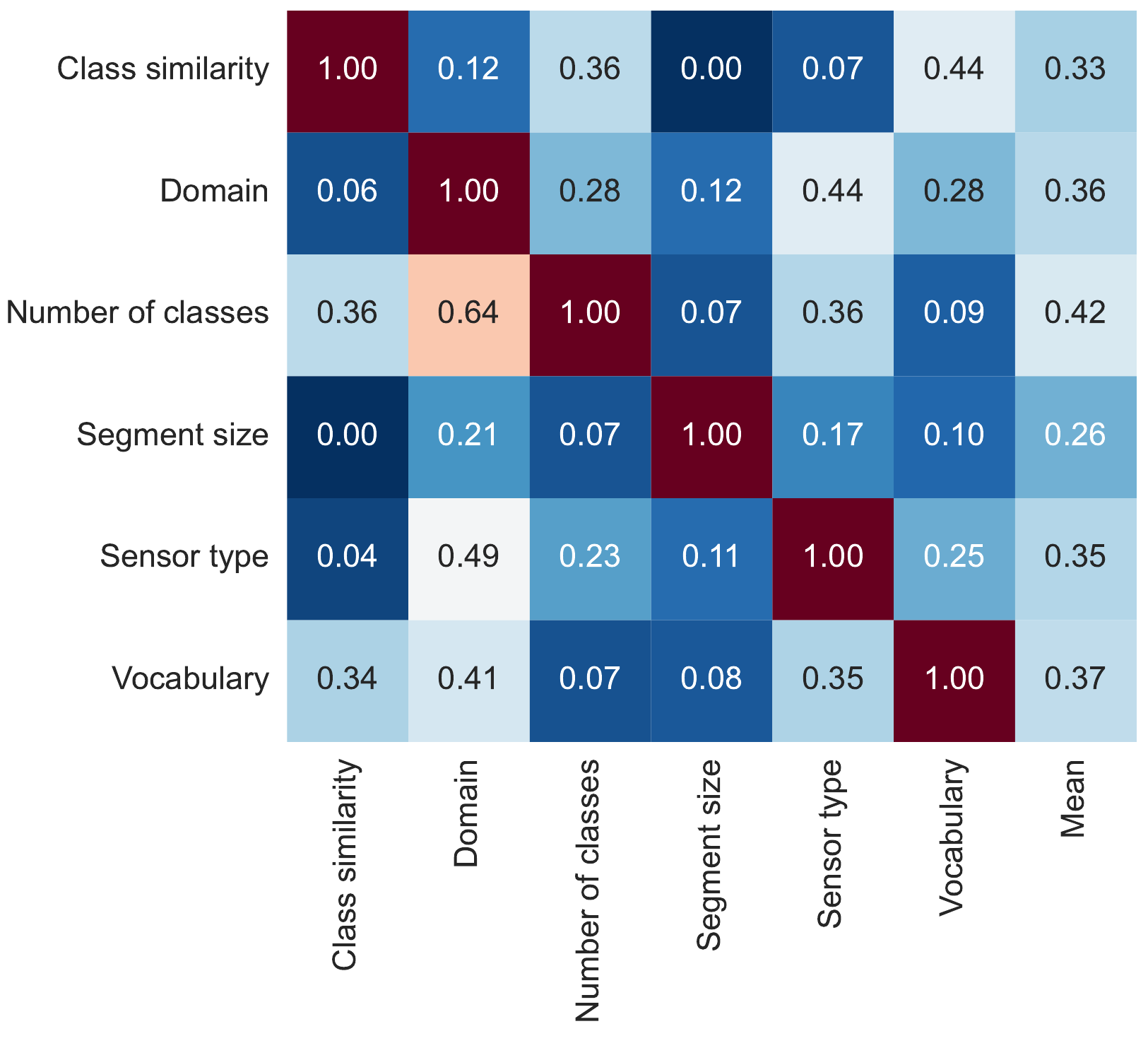} }}
    \hspace{1cm}
  
  \caption{Peason correlation between dimensions based on the classified datasets for an interim status~(a) and the final taxonomy~(b).}\label{fig:correlation}
\end{figure}

Finally, we utilized a statistical analysis to identify similar dimensions, specifically, the Pearson correlation between each pair of dimensions using the empirical data from the classified datasets. We applied one-hot encoding for categorical dimensions and scaled each ordinal dimension by the number of characteristics. Figure~\ref{fig:correlation} visualize multiple pairs with high correlation, e.g., \textit{segment size} and \textit{number of segments}. We reduced the interim dimensions based on the statistical analysis and the meta-characteristic. The final taxonomy passes all ending conditions in~\cite{TaxonomyNickerson2013}.

\begin{tiny}
\setlength\extrarowheight{0.2em}
\addtolength{\tabcolsep}{-0.3em}

\begin{xltabular}{\textwidth}{lc|ccYYYc}

\caption{All 120 classified semantic segmentation datasets within the taxonomy.}\label{tab:classification} \\

\toprule
Dataset & Task & Domain & Sensor type & Segment size & Number of classes & Class similarity & Vocabulary\\ 
\midrule
\endfirsthead
\toprule
Dataset & Task & Domain & Sensor type & Segment size & Number of classes & Class similarity & Vocabulary\\ 
\midrule
\endhead
\bottomrule
\endfoot
\bottomrule
\endlastfoot
COCO Stuff \cite{DatasetCOCO} & Common & \multirow{36}{*}{General} & Visible spectrum & Medium & 171 (Many) & Low & Generic \\
Pascal VOC 2012 \cite{DatasetPascalVOC} & Common &  & Visible spectrum & Medium & 20 (Medium) & Low & Generic \\
ADE20K-150 \cite{DatasetADE20K} & Common &  & Visible spectrum & Medium & 150 (Many) & Low & Generic \\
ADE20K-847 \cite{DatasetADE20K} & Common &  & Visible spectrum & Medium & 847 (Many) & High & Generic \\
Pascal Context-59 \cite{DatasetPascalContext} & Common &  & Visible spectrum & Medium & 59 (Many) & Low & Generic \\
Pascal Context-459 \cite{DatasetPascalContext} & Common &  & Visible spectrum & Medium & 459 (Many) & High & Generic \\
LVIS \cite{DatasetLVIS} & Common &  & Visible spectrum & Small & 1203 (Many) & High & Generic \\
FSS-1000 \cite{DatasetFSS1000} & Common &  & Visible spectrum & Large & 1000 (Many) & High & Generic \\
Mapillary Vistas v1 \cite{DatasetMapillaryVistas} & Driving &  & Visible spectrum & Small & 66 (Many) & Low & Generic \\
Mapillary Vistas v2 \cite{DatasetMapillaryVistas} & Driving &  & Visible spectrum & Small & 124 (Many) & Low & Task-spec. \\
Cityscapes \cite{DatasetCityscapes} & Driving &  & Visible spectrum & Small & 30 (Medium) & Low & Generic \\
BDD100K \cite{DatasetBDD100K} & Driving &  & Visible spectrum & Medium & 19 (Medium) & Low & Generic \\
Dark Zurich \cite{DatasetDarkZurich} & Driving &  & Visible spectrum & Medium & 20 (Medium) & Low & Generic \\
SYNTHIA \cite{DatasetSYNTHIA} & Driving &  & Visible spectrum & Small & 13 (Medium) & Low & Generic \\
WoodScape  \cite{DatasetWoodScape} & Driving &  & Visible spectrum & Small & 40 (Medium) & High & Generic \\
MVTec D2S \cite{DatasetMVTecD2S} & Checkout &  & Visible spectrum & Medium & 60 (Many) & Low & Generic \\
EgoHands \cite{DatasetEgoHands} & Ego hands &  & Visible spectrum & Medium & 5 (Few) & High & Task-spec. \\
WorkingHands \cite{DatasetWorkingHands} & Ego hands &  & Visible spectrum & Medium & 16 (Medium) & Low & Generic \\
EgoHOS \cite{DatasetEgoHOS} & Ego hands &  & Visible spectrum & Medium & 8 (Few) & High & Task-spec. \\
EYTH \cite{DatasetEYTH} & Ego hands &  & Visible spectrum & Medium & 2 (Binary) & Low & Generic \\
VISOR \cite{DatasetVISOR} & Ego hands &  & Visible spectrum & Small & 257 (Many) & High & Generic \\
Open Surfaces \cite{DatasetOpenSurfaces} & Materials &  & Visible spectrum & Medium & 37 (Medium) & High & Domain-spec. \\
MINC \cite{DatasetMINC} & Materials &  & Visible spectrum & Medium & 23 (Medium) & Low & Generic \\
DMS \cite{DatasetDMS} & Materials &  & Visible spectrum & Small & 52 (Many) & High & Generic \\
DeepFashion2 \cite{DatasetDeepFashion2} & Clothing &  & Visible spectrum & Small & 13 (Medium) & Low & Generic \\
ModaNet \cite{DatasetModaNet} & Clothing &  & Visible spectrum & Small & 13 (Medium) & Low & Generic \\
MHP v1 \cite{DatasetMHP} & Body parts &  & Visible spectrum & Small & 18 (Medium) & High & Task-spec. \\
MHP v2 \cite{DatasetMHP} & Body parts &  & Visible spectrum & Small & 58 (Many) & High & Task-spec. \\
FoodSeg103 \cite{DatasetFoodSeg103} & Ingredients &  & Visible spectrum & Medium & 103 (Many) & High & Generic \\
TACO \cite{DatasetTACO} & Trash &  & Visible spectrum & Medium & 60 (Many) & High & Domain-spec. \\
RailSem19  \cite{DatasetRailSem19} & Rail &  & Visible spectrum & Small & 11 (Medium) & High & Task-spec. \\
ATLANTIS \cite{DatasetATLANTIS} & Maritime &  & Visible spectrum & Small & 56 (Many) & Low & Generic \\
Aircraft Context \cite{DatasetAircraftContext} & Aerial vehicles &  & Visible spectrum & Medium & 8 (Few) & Low & Generic \\
RELLIS-3D \cite{DatasetRELLIS3D} & Robotics &  & Visible spectrum & Small & 20 (Medium) & Low & Generic \\
SketchyScene-7k \cite{DatasetSketchyScene7k} & Sketches &  & Visible spectrum & Small & 45 (Medium) & Low & Generic \\
DRAM \cite{DatasetDRAM} & Paintings &  & Visible spectrum & Medium & 12 (Medium) & Low & Generic \\
\hline
iSAID \cite{DatasetiSAID} & Objects & \multirow{27}{*}{\shortstack{Earth\\Monitoring}} & Visible spectrum & Small & 15 (Medium) & Low & Generic \\
DSTL Satellite \cite{DatasetDSTLSatellite} & Objects &  & Multispectral & Small & 10 (Medium) & High & Generic \\
ISPRS Potsdam \cite{DatasetISPRSPotsdam} & Land use &  & Multispectral & Small & 6 (Few) & Low & Generic \\
LandCoverNet \cite{DatasetLandCoverNet} & Land use &  & Multispectral & Medium & 7 (Few) & Low & Generic \\
LoveDA \cite{DatasetLoveDA} & Land use &  & Visible spectrum & Small & 7 (Few) & Low & Generic \\
Deep Globe \cite{DatasetDeepGlobe} & Land use &  & Visible spectrum & Medium & 7 (Few) & Low & Generic \\
GID-5 \cite{DatasetGID} & Land use &  & Multispectral & Small & 5 (Few) & Low & Generic \\
GID-15 \cite{DatasetGID} & Land use &  & Multispectral & Small & 16 (Medium) & High & Task-spec. \\
Dubai \cite{DatasetDubai} & Land use &  & Visible spectrum & Small & 6 (Few) & Low & Generic \\
Sen1Floods11 \cite{DatasetSen1Floods11} & Floods &  & Electromagnetic & Small & 2 (Binary) & Low & Generic \\
WorldFloods \cite{DatasetWorldFloods} & Floods &  & Multispectral & Medium & 3 (Binary) & Low & Generic \\
HR-GLDD \cite{DatasetHRGLDD} & Landslides &  & Multispectral & Medium & 2 (Binary) & Low & Generic \\
Antarctic fracture \cite{DatasetAntarcticfracture} & Ice fractures &  & Multispectral & Small & 2 (Binary) & Low & Generic \\
Active fire \cite{DatasetActivefire} & Wildfires &  & Multispectral & Small & 2 (Binary) & Low & Generic \\
xBD \cite{DatasetxBD} & Buildings &  & Visible spectrum & Small & 5 (Few) & High & Task-spec. \\
MSAW \cite{DatasetMSAW} & Buildings &  & Electromagnetic & Small & 2 (Binary) & Low & Generic \\
3D PV Locator \cite{Dataset3DPVLocator} & PV &  & Visible spectrum & Small & 2 (Binary) & Low & Generic \\
AgricultureVision \cite{DatasetAgricultureVision} & Agriculture &  & Multispectral & Medium & 9 (Few) & Low & Domain-spec. \\
PASTIS \cite{DatasetPASTIS} & Agriculture &  & Multispectral & Small & 18 (Medium) & High & Domain-spec. \\
CalCROP21 \cite{DatasetCalCROP21} & Agriculture &  & Multispectral & Small & 29 (Medium) & High & Domain-spec. \\
Arctic Sea Ice \cite{DatasetArcticSeaIce} & Sea ice &  & Multispectral & Medium & 8 (Few) & High & Task-spec. \\
ELAI Dust Storm \cite{DatasetELAIDustStorm} & Dust storm &  & Visible spectrum & Large & 2 (Binary) & Low & Generic \\
FloodNet \cite{DatasetFloodNet} & Floods &  & Visible spectrum & Medium & 10 (Few) & Low & Task-spec. \\
SDD \cite{DatasetSDD} & Objects &  & Visible spectrum & Small & 21 (Medium) & Low & Generic \\
UDD \cite{DatasetUDD} & Objects &  & Visible spectrum & Medium & 6 (Few) & Low & Generic \\
UAVid \cite{DatasetUAVid} & Objects &  & Visible spectrum & Small & 6 (Few) & High & Task-spec. \\
PV thermography \cite{DatasetPVthermography} & PV &  & Electromagnetic & Small & 6 (Binary) & High & Domain-spec. \\
\hline
CholecSeg8k \cite{DatasetCholecSeg8k} & Surgery & \multirow{10}{*}{\shortstack{Medical\\Sciences}} & Visible spectrum & Medium & 13 (Medium) & High & Domain-spec. \\
RoboTool \cite{DatasetRoboTool} & Surgery &  & Visible spectrum & Medium & 2 (Binary) & Low & Generic \\
Kvasir-Instrument \cite{DatasetKvasirInstrument} & Surgery &  & Visible spectrum & Medium & 2 (Binary) & Low & Generic \\
ROBUST-MIS 2019 \cite{DatasetROBUSTMIS2019} & Surgery &  & Visible spectrum & Medium & 2 (Binary) & Low & Generic \\
Kvasir SEG \cite{DatasetKvasirSEG} & Surgery &  & Visible spectrum & Medium & 2 (Binary) & Low & Domain-spec. \\
Vocalfolds \cite{DatasetVocalfolds} & Surgery &  & Visible spectrum & Medium & 7 (Few) & Low & Domain-spec. \\
CHASE DB1 \cite{DatasetCHASEDB1} & Retina scan &  & Microscopic & Small & 2 (Binary) & Low & Domain-spec. \\
HRF \cite{DatasetHRF} & Retina scan &  & Microscopic & Small & 2 (Binary) & Low & Domain-spec. \\
STARE \cite{DatasetSTARE} & Retina scan &  & Microscopic & Small & 2 (Binary) & Low & Domain-spec. \\
Intraretinal C. Fluid \cite{DatasetIntraretinalCFluid} & Retinal OCT &  & Microscopic & Small & 2 (Binary) & Low & Domain-spec. \\
GLaS \cite{DatasetGLaS} & WSI & \multirow{14}{*}{\shortstack{Medical\\Sciences}} & Microscopic & Medium & 2 (Binary) & High & Domain-spec. \\
Gleason \cite{DatasetGleason} & WSI &  & Microscopic & Large & 6 (Few) & High & Domain-spec. \\
CryoNuSeg \cite{DatasetCryoNuSeg} & WSI &  & Microscopic & Small & 2 (Binary) & Low & Domain-spec. \\
BBBC038v1 \cite{DatasetBBBC038v1} & WSI &  & Microscopic & Small & 2 (Binary) & Low & Domain-spec. \\
Vector-LabPics \cite{DatasetVectorLabPics} & Lab vessels &  & Visible spectrum & Medium & 58 (Medium) & High & Domain-spec. \\
vesselNN \cite{DatasetvesselNN} & Brain vessel &  & Microscopic & Small & 2 (Binary) & Low & Domain-spec. \\
MTNeuro  \cite{DatasetMTNeuro} & Brain vessel &  & Microscopic & Small & 3 (Few) & High & Domain-spec. \\
Neuronal Cells \cite{DatasetNeuronalCells} & Brain cells &  & Microscopic & Small & 2 (Binary) & Low & Domain-spec. \\
BraTS 2015 \cite{DatasetBraTS2015} & Brain tumor &  & Electromagnetic & Medium & 5 (Few) & High & Domain-spec. \\
ISIC2018 Task1 \cite{DatasetISIC2018Task1} & Lesions &  & Visible spectrum & Large & 2 (Binary) & Low & Domain-spec. \\
PAXRay-166 \cite{DatasetPAXRay} & X-Ray &  & Electromagnetic & Small & 166x2 (Binary) & High & Domain-spec. \\
PAXRay-4 \cite{DatasetPAXRay} & X-Ray &  & Electromagnetic & Large & 4x2 (Binary) & Low & Domain-spec. \\
Pulmonary Chest \cite{DatasetPulmonaryChest} & X-Ray &  & Electromagnetic & Large & 2 (Binary) & Low & Generic \\
US segmentation \cite{DatasetUSsegmentation} & Ultrasound &  & Electromagnetic & Medium & 9 (Few) & High & Domain-spec. \\
\hline
Severstal \cite{DatasetSeverstal} & Surface defect & \multirow{15}{*}{Engineering} & Visible spectrum & Medium & 4 (Few) & High & Domain-spec. \\
KolektorSDD2 \cite{DatasetKolektorSDD2} & Surface defect &  & Visible spectrum & Medium & 2 (Binary) & Low & Generic \\
EMPS \cite{DatasetEMPS} & Particles &  & Electromagnetic & Small & 2 (Binary) & Low & Generic \\
LIB-HSI \cite{DatasetLIBHSI} & Building fasade &  & Multispectral & Medium & 44 (Medium) & High & Generic \\
Corrosion CS \cite{DatasetCorrosionCS} & Corrosion &  & Visible spectrum & Medium & 4 (Few) & High & Task-spec. \\
LCW \cite{DatasetLCW} & Cracks &  & Visible spectrum & Small & 2 (Binary) & Low & Generic \\
DeepCrack \cite{DatasetDeepCrack} & Cracks &  & Visible spectrum & Small & 2 (Binary) & Low & Generic \\
ZeroWaste-f \cite{DatasetZeroWaste} & Conveyor &  & Visible spectrum & Medium & 4 (Few) & High & Generic \\
Thermal Dog \cite{DatasetThermalDog} & Thermal &  & Electromagnetic & Medium & 3 (Few) & Low & Generic \\
PST900 \cite{DatasetPST900} & Thermal &  & Electromagnetic & Small & 5 (Few) & Low & Generic \\
TAS-NIR \cite{DatasetTASNIR} & Thermal &  & Electromagnetic & Medium & 22 (Medium) & High & Generic \\
PIDRay \cite{DatasetPIDRay} & Security &  & Electromagnetic & Small & 12 (Medium) & Low & Generic \\
TTPLA \cite{DatasetTTPLA} & Powerlines &  & Visible spectrum & Small & 5 (Few) & High & Generic \\
Vale \cite{DatasetVale} & Terrain &  & Visible spectrum & Medium & 5 (Few) & High & Task-spec. \\
AI4MARS \cite{DatasetAI4MARS} & Terrain &  & Visible spectrum & Small & 4 (Few) & High & Generic \\
\hline
TrashCan \cite{DatasetTrashCan} & Trash & \multirow{18}{*}{\shortstack{Agriculture\\and Biology}} & Visible spectrum & Medium & 4 (Few) & Low & Generic \\
SUIM \cite{DatasetSUIM} & Underwater &  & Visible spectrum & Medium & 8 (Few) & Low & Generic \\
DeepFish \cite{DatasetDeepFish} & Fish &  & Visible spectrum & Medium & 2 (Binary) & Low & Generic \\
NDD20 \cite{DatasetNDD20} & Fish &  & Visible spectrum & Medium & 2 (Binary) & Low & Generic \\
Ciona17 \cite{DatasetCiona17} & Maritime species &  & Visible spectrum & Large & 4 (Few) & High & Domain-spec. \\
CUB-200 \cite{DatasetCUB200} & Bird species &  & Visible spectrum & Medium & 201 (Many) & High & Domain-spec. \\
Oxford-IIIT Pet \cite{DatasetOxfordIIITPet} & Animal species &  & Visible spectrum & Large & 28 (Medium) & High & Domain-spec. \\
Plittersdorf \cite{DatasetPlittersdorf} & Animals &  & Electromagnetic & Medium & 2 (Binary) & Low & Generic \\
CAMO \cite{DatasetCAMO} & Animals &  & Visible spectrum & Medium & 2 (Medium) & Low & Domain-spec. \\
COD \cite{DatasetCOD} & Animals &  & Visible spectrum & Medium & 78 (Many) & Low & Domain-spec. \\
CropAndWeed \cite{DatasetCropAndWeed} & Plants &  & Visible spectrum & Small & 100 (Many) & High & Domain-spec. \\
WGISD \cite{DatasetWGISD} & Plants &  & Visible spectrum & Medium & 2 (Binary) & Low & Generic \\
PPDPS \cite{DatasetPPDPS} & Plants &  & Visible spectrum & Large & 2 (Binary) & Low & Generic \\
Plant seg. \cite{DatasetPlantseg} & Plants &  & Visible spectrum & Small & 3 (Few) & High & Task-spec. \\
CWFID \cite{DatasetCWFID} & Crops &  & Visible spectrum & Small & 3 (Few) & High & Generic \\
PPDLS \cite{DatasetPPDPS} & Leefs &  & Visible spectrum & Medium & 2 (Binary) & Low & Generic \\
Leaf disease \cite{DatasetLeafdisease} & Leef disease &  & Visible spectrum & Small & 2 (Binary) & Low & Generic \\
Rice Leaf dis. \cite{DatasetRiceLeafdis} & Leef disease &  & Visible spectrum & Small & 5 (Few) & High & Domain-spec. \\

\end{xltabular}
\end{tiny}

\section{Benchmark datasets}\label{sec:datasets}

\subsection{Overview}
We selected 22 out of the 120 classified datasets for the MESS benchmark. The links, licenses, selected splits, and a sample of the class labels of the datasets are provided in Table \ref{tab:details}. We specified some label names for better performances of the models similar to~\cite{GLIP2022}. E.g., we use \textit{crop seedling} instead of \textit{crop} for the CWFID dataset. We refer to our implementation for all class labels.
 
\begin{table}[tbh]
    \caption{Details for the 22 MESS datasets including the links and licenses. Nearly all datasets require attribution and many only allow non-commercial use.}
    \label{tab:details}
    
    \begin{center}
    \tiny
    \setlength\extrarowheight{0.2em}
    \addtolength{\tabcolsep}{-0.3em}
    \begin{minipage}{\textwidth}
    \begin{tabularx}{\textwidth}{lllYYl}
\toprule
Dataset & Link & Licence & Split & No. of classes & Classes \\
\midrule
BDD100K \cite{DatasetBDD100K} & \href{https://bdd-data.berkeley.edu}{berkeley.edu} & \href{https://doc.bdd100k.com/license.html}{custom} & val & 19 & [road; sidewalk; building; wall; fence; pole; traffic light; traffic sign; ...] \\
Dark Zurich \cite{DatasetDarkZurich} & \href{https://www.trace.ethz.ch/publications/2019/GCMA_UIoU/}{ethz.ch} & custom & val & 20 & [unlabeled; road; sidewalk; building; wall; fence; pole; traffic light; ...] \\
MHP v1 \cite{DatasetMHP} & \href{https://github.com/ZhaoJ9014/Multi-Human-Parsing}{github.com} & \href{https://lv-mhp.github.io/}{custom} & test & 19 & [others; hat; hair; sunglasses; upper clothes; skirt; pants; dress; ...] \\
FoodSeg103 \cite{DatasetFoodSeg103} & \href{https://xiongweiwu.github.io/foodseg103.html}{github.io} & Apache 2.0 & test & 104 & [background; candy; egg tart; french fries; chocolate; biscuit; popcorn; ...] \\
ATLANTIS \cite{DatasetATLANTIS} & \href{https://github.com/smhassanerfani/atlantis}{github.com} & Flickr (images) & test & 56 & [bicycle; boat; breakwater; bridge; building; bus; canal; car; ...] \\
DRAM \cite{DatasetDRAM} & \href{https://faculty.runi.ac.il/arik/site/artseg/Dram-Dataset.html}{ac.il} & custom (in download) & test & 12 & [bird; boat; bottle; cat; chair; cow; dog; horse; ...] \\
\hline
iSAID \cite{DatasetiSAID} & \href{https://captain-whu.github.io/iSAID/dataset.html}{github.io} & Google Earth (images) & val & 16 & [others; boat; storage tank; baseball diamond; tennis court; bridge; ...] \\
ISPRS Potsdam \cite{DatasetISPRSPotsdam} & \href{https://www.isprs.org/education/benchmarks/UrbanSemLab/default.aspx}{isprs.org} & no licence provided\footnote{Upon request, the naming of the data provider and project is required.} & test & 6 & [road; building; grass; tree; car; others] \\
WorldFloods \cite{DatasetWorldFloods} & \href{https://github.com/spaceml-org/ml4floods/blob/main/jupyterbook/content/worldfloods_dataset.md}{github.com} & CC NC 4.0 & test & 3 & [land; water and flood; cloud] \\
FloodNet \cite{DatasetFloodNet} & \href{https://github.com/BinaLab/FloodNet-Supervised_v1.0}{github.com} & \href{https://cdla.dev/permissive-1-0/}{custom} & test & 10 & [building-flooded; building-non-flooded; road-flooded; water; tree; ...] \\
UAVid \cite{DatasetUAVid} & \href{https://uavid.nl}{uavid.nl} & CC BY-NC-SA 4.0 & val & 8 & [others; building; road; tree; grass; moving car; parked car; humans] \\
\hline
Kvasir-Inst. \cite{DatasetKvasirInstrument} & \href{https://datasets.simula.no/kvasir-instrument/}{simula.no} & \href{https://datasets.simula.no/kvasir-instrument/}{custom} & test & 2 & [others; tool] \\
CHASE DB1 \cite{DatasetCHASEDB1} & \href{https://blogs.kingston.ac.uk/retinal/chasedb1/}{kingston.ac.uk} & CC BY 4.0 & test & 2 & [others; blood vessels] \\
CryoNuSeg \cite{DatasetCryoNuSeg} & \href{https://www.kaggle.com/datasets/ipateam/segmentation-of-nuclei-in-cryosectioned-he-images}{kaggle.com} & CC BY-NC-SA 4.0 & test & 2 & [others; nuclei in cells] \\
PAXRay-4 \cite{DatasetPAXRay} & \href{https://constantinseibold.github.io/paxray/}{github.io} & \href{https://constantinseibold.github.io/paxray/}{custom} & test & 4x2 & [others, lungs], [others, bones], [others, mediastinum], [others, diaphragm] \\
\hline
Corrosion CS \cite{DatasetCorrosionCS} & \href{https://figshare.com/articles/dataset/Corrosion_Condition_State_Semantic_Segmentation_Dataset/16624663}{figshare.com} & CC0 & test & 4 & [others; steel with fair corrosion; ... poor corrosion; ... severe corrosion] \\
DeepCrack \cite{DatasetDeepCrack} & \href{https://github.com/yhlleo/DeepCrack/tree/master}{github.com} & \href{https://github.com/yhlleo/DeepCrack/tree/master}{custom} & test & 2 & [concrete or asphalt; crack] \\
PST900 \cite{DatasetPST900} & \href{https://github.com/ShreyasSkandanS/pst900_thermal_rgb}{github.com} & GPL-3.0 & test & 5 & [background; fire extinguisher; backpack; drill; human] \\
ZeroWaste-f \cite{DatasetZeroWaste} & \href{http://ai.bu.edu/zerowaste/}{ai.bu.edu} & CC-BY-NC 4.0 & test & 5 & [background or trash; rigid plastic; cardboard; metal; soft plastic] \\
\hline
SUIM \cite{DatasetSUIM} & \href{https://irvlab.cs.umn.edu/resources/suim-dataset}{umn.edu} & MIT & test & 8 & [human diver; reefs and invertebrates; fish and vertebrates; ...] \\
CUB-200 \cite{DatasetCUB200} & \href{https://www.vision.caltech.edu/datasets/cub_200_2011/}{caltech.edu} & \href{https://www.vision.caltech.edu/datasets/cub_200_2011/}{custom} & test & 201 & [background; Laysan Albatross; Sooty Albatross; Crested Auklet; ...] \\
CWFID \cite{DatasetCWFID} & \href{https://github.com/cwfid/dataset}{github.com} & \href{https://github.com/cwfid/dataset}{custom} & test & 3 & [ground; crop seedling; weed] \\
\bottomrule
\end{tabularx}

    \end{minipage}
    \end{center}
\end{table}


We shortly introduce each dataset in the following: The general datasets include datasets with everyday scenes but more specific use cases and niche image themes in comparison to the standard evaluation datasets. Specifically, the use cases include two driving datasets with one covering nighttime images. Further, MHP v1 covers classes of body parts and clothes while FoodSeg103 requires the segmentation of different ingredients. The ATLANTIS dataset focuses on classes related to maritime environments and DRAM covers common classes in paintings.
The selected earth monitoring datasets include iSAID, which requires the segmentation of 15 object categories in satellite images, e.g., a tennis court or a helicopter. ISPRS Potsdam and WorldFloods provide multispectral data, and our main evaluation uses an IRRG false color mapping. Near-infrared radiation is visualized in red and highlights vegetation. ISPRS Potsdam provides very high-resolution images of an urban area with multiple classes, while WorldFloods has a 10-meter resolution and focuses on water segmentation. We selected two drone datasets with similar use cases. UAVid includes urban scenes, and FloodNet covers flooded buildings and roads. 
The medical datasets cover four different modalities: Endoscopy (RGB images), retinal scans, whole slide imagery (WSI), and X-ray scans. Each binary segmentation task focuses on a specific object or anatomical structure, like blood vessels or lungs. The multi-label segmentation dataset PAXRay is a special case. We do not use each of the 166 annotated classes but only the four superclasses. Because of the mask overlay, each class is predicted in a binary setting, and we average the resulting metrics.
Next, we selected four diverse engineering datasets. Corrosion~CS includes images of corrosion on bridges and other infrastructure with four different condition states. DeepCrack consists of close-up images of crack. PST900 consists of thermal imagery with firefighter-related objects. We use a gray-scale color map in our main evaluation to visualize the thermal data. The Zero-Waste-f dataset includes images of a conveyor belt with annotations for four types of recyclable trash.
The final three datasets cover biological-related datasets: SUIM is an underwater imagery dataset with fish, aquatic plants, and others. CUB-200 is a widely used dataset of 200 bird species. The images of CUB-200 are relatively easy to segment, but assigning the correct species is challenging. CWFID includes crop seedlings and weeds.

We looked up the current fully supervised performance to provide an upper threshold for each dataset and present them in Table \ref{tab:datasets_supervised}.
We did not find any mIoU results for the MHP v1 dataset as it is originally annotated for instance segmentation. Therefore, we trained MaskFormer~\cite{MaskFormer2021} to provide a reference. We trained the model for 100K steps using the Swin-B ADE20K-150 settings and evaluated the best model based on the val mIoU.

\begin{table}[htb]
\caption{Supervised mIoU results for the datasets.}\label{tab:datasets_supervised}

\begin{center}
\small
\begin{minipage}{\textwidth}
\centering
\begin{tabular}{llcc}
\toprule
Dataset & Model & Year & mIoU \\ 
\midrule
BDD100K & Two-branch Enet \cite{sotaBDD100K} & 2023 & 44.8 \\
Dark Zurich & Refign (HRDA) \cite{sotaDarkZurich} & 2023 & 63.9 \\
MHP v1 & MaskFormer (Swin-B) \cite{MaskFormer2021} & 2021 & 53.18\footnote{Own experiment because mIoU results are not reported in MHP v1 literature.} \\
FoodSeg103 & SeTR-MLA (ViT-16/B) \cite{sotaFoodSeg103} & 2021 & 45.1 \\
ATLANTIS & AQUANet \cite{DatasetATLANTIS} & 2021 & 42.22 \\
DRAM & DRAM model \cite{DatasetDRAM} & 2022 & 45.71\footnote{The DRAM model is not trained on a labeled training set but self-supervised on generated images.} \\
iSAID & IMP-ViTAEv2-S-UperNet \cite{sotaiSAID} & 2022 & 65.3 \\
ISPRS Potsdam & DC-Swin \cite{sotaISPRSPotsdam} & 2022 & 87.56 \\
WorldFloods & UNet \cite{DatasetWorldFloods} & 2021 & 92.71 \\
FloodNet & SegFormer \cite{sotaFloodNet} & 2023 & 82.22 \\
UAVid & UNetFormer \cite{sotaUAVid} & 2022 & 67.8 \\
Kvasir-Instrument & U-Net \cite{DatasetKvasirInstrument} & 2021 & 93.7 \\
CHASE DB1 & RV-GAN \cite{sotaCHASEDB1} & 2021 & 97.05 \\
CryoNuSeg & TransUNet \cite{sotaCryoNuSeg} & 2022 & 73.45 \\
PAXRay-4 & Unet-R50 \cite{DatasetPAXRay} & 2022 & 93.77 \\
Corrosion CS & DeeplabV3+ \cite{DatasetCorrosionCS} & 2021 & 49.92 \\
DeepCrack & DeepCrack-GF \cite{DatasetDeepCrack} & 2019 & 85.9 \\
ZeroWaste-f & DeeplabV3+ \cite{DatasetZeroWaste} & 2022 & 52.5 \\
PST900 & SpiderMesh \cite{sotaPST900} & 2022 & 82.3 \\
SUIM & LOCA \cite{sotaSUIM} & 2022 & 74.0 \\
CUB-200 & GFN \cite{sotaCUB200} & 2022 & 84.6 \\
CWFID & Unet-Resnet-50 \cite{sotaCWFID} & 2022 & 87.23 \\
\bottomrule
\end{tabular}
\end{minipage}
\end{center}

\end{table}

\subsection{Dataset analysis}

\begin{figure}[tbh]
    \centering
    \subfloat[\centering PCA]{{\includegraphics[width=0.49\textwidth]{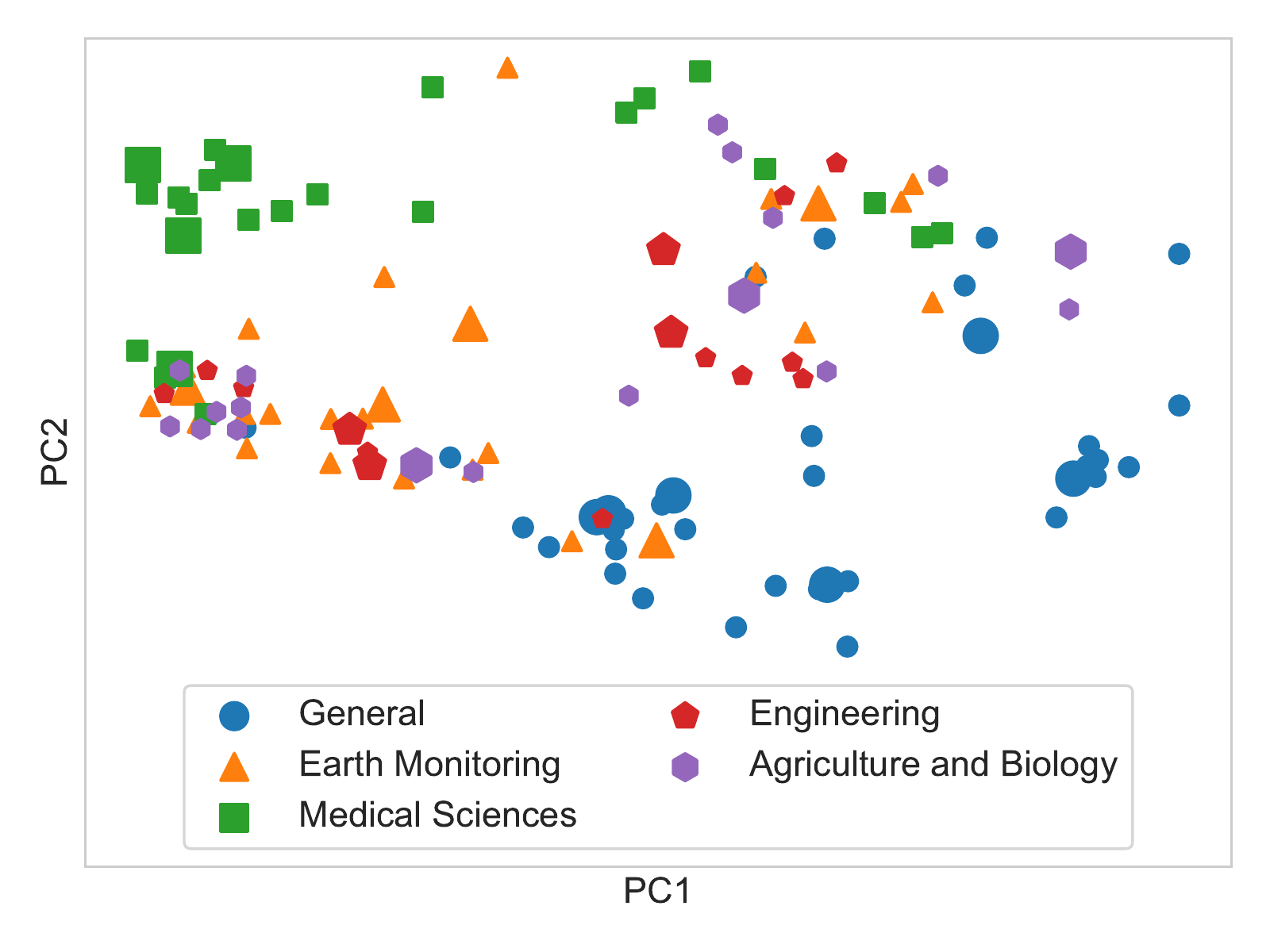} }}
    \subfloat[\centering Influence]{{\includegraphics[width=0.49\textwidth]{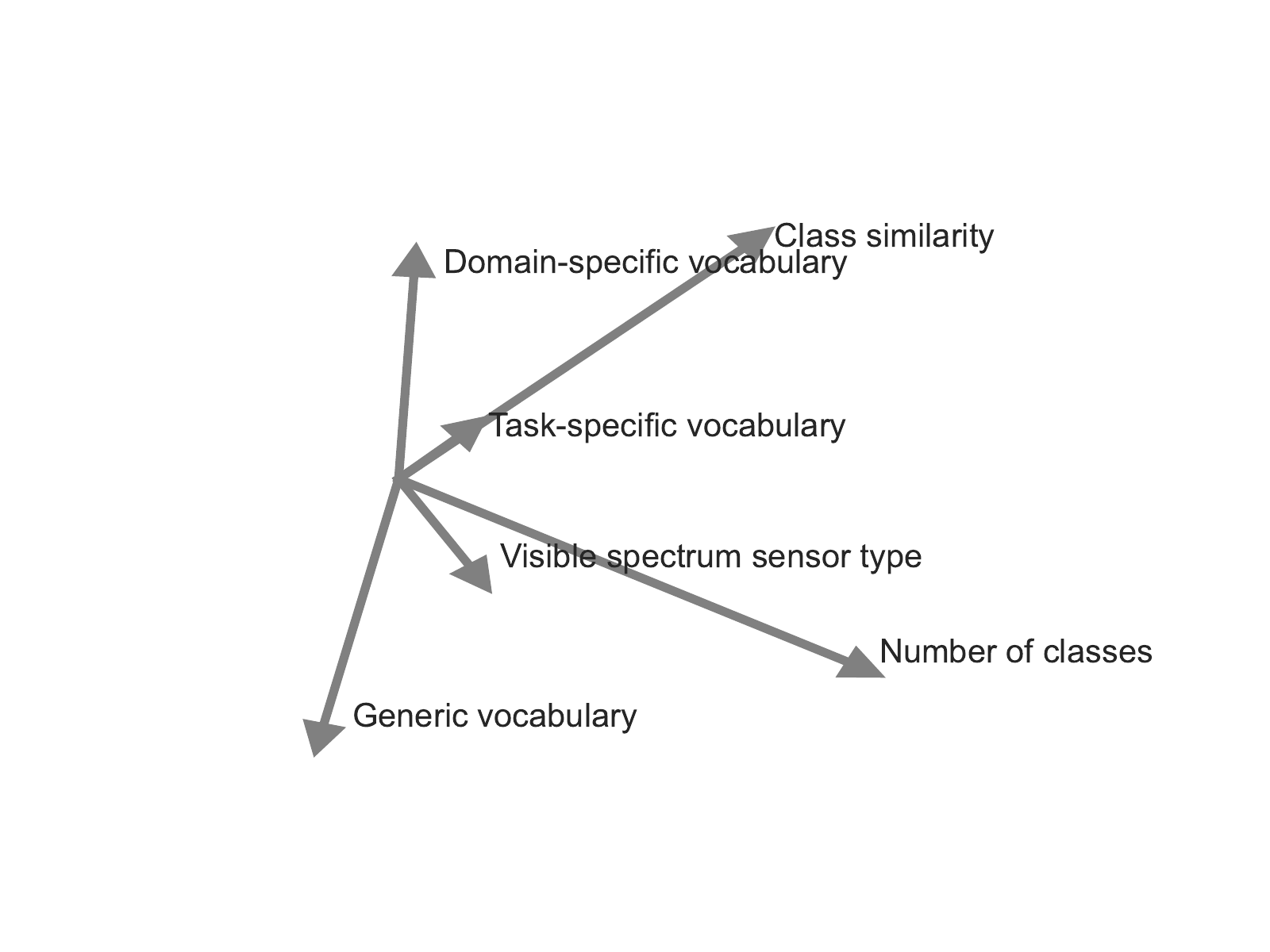} }}
    
  \caption[Principle Component Analysis of the classified datasets]{PCA of the classified datasets, clustered by their domain (a), and the highest influencing factors apart from the domains (b). An increased size visualizes selected datasets. Some noise was added to visualize similar classified datasets.}\label{fig:pca}
\end{figure}

The classified datasets are visualized in Figure \ref{fig:pca} by applying a Principal Component Analysis (PCA) along the taxonomy’s dimensions. An analysis of the principal components reveals that, apart from the domain, mainly language-related features differentiate the datasets within the taxonomy. The PCA has two big clusters covering all domains – one cluster of datasets (top) with mostly domain-specific vocabulary and high class similarity and another one (bottom) with tasks of easily distinguishable generic classes. The PCA emphasizes the importance of these two dimensions for all domains. The datasets visualized in the center between these clusters have either a domain-specific vocabulary with low class similarly, which is often the case for medical datasets, or the opposite, often observed in general datasets.
Furthermore, medical datasets have few classes, while general use cases have many classes, with the other three domains in between.

Following \citet{SAN2023}, we conducted an analysis of the similarity between the labels of each dataset and the training labels from COCO-Stuff~\cite{DatasetCOCO} which is used by most evaluated models. The similarity between two labels is computed using the cosine similarity between their CLIP text embeddings. Next, we select the maximum similarity value for each text label (i.e., the minimal distance of this label to the training labels).
To calculate the similarity between a test set and the training set, we can select the minimum value among the test labels. This represents a Hausdoff Distance between these two sets, i.e., the maximum distance in the embedding space~\cite{SAN2023}. However, this calculation is sensitive to outliers and we also report the mean similarity over all test labels. 

\begin{figure}[tbh]
    \includegraphics[width=0.85\textwidth]{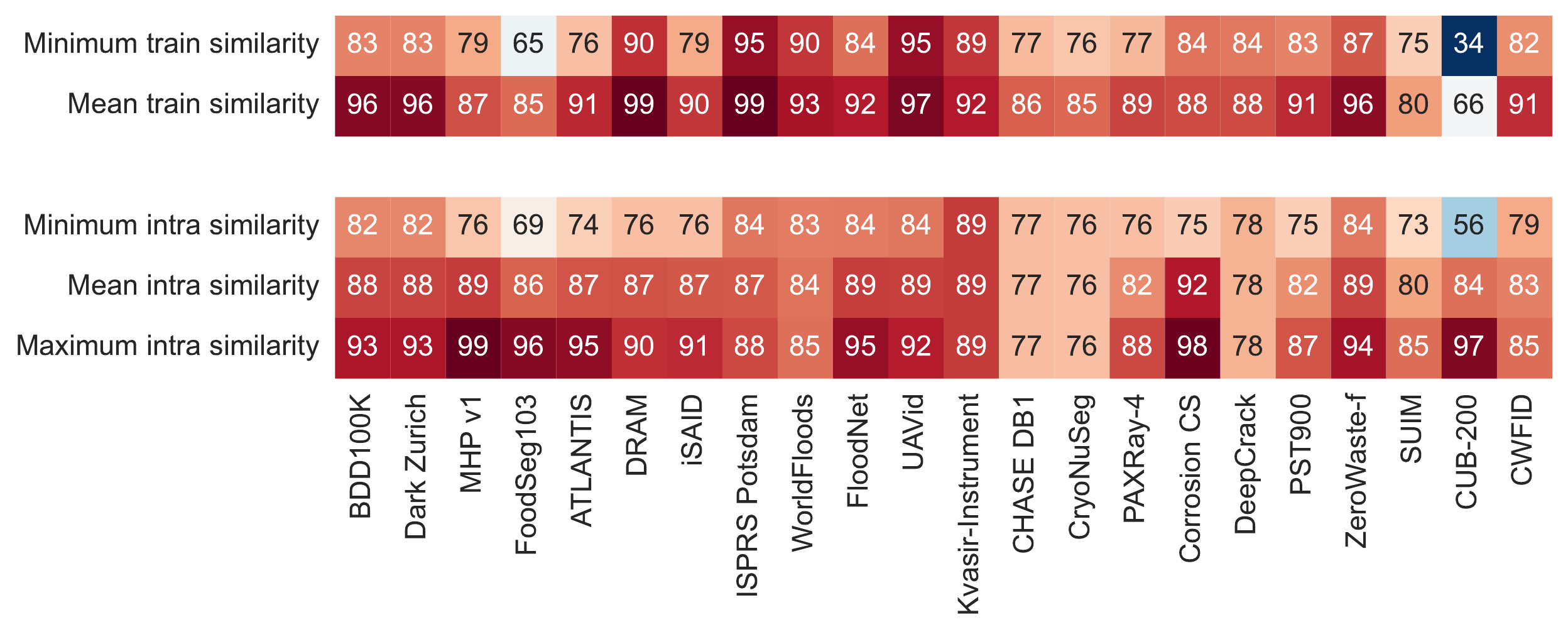}
    
    \caption{Class similarity to the COCO-Stuff training labels and within each dataset.}
    \label{fig:dataset_corr}
\end{figure}

The analysis in Figure~\ref{fig:dataset_corr} visualizes that most datasets do include classes with a low train similarity that are not related to the train labels. Some datasets have a high mean similarity (i.a., BDD100K, DRAM, ISPRS Potsdam, ZeroWaste-f). Therefore, most classes in these datasets are equal or similar to a training label from COCO-Stuff. 
The medical and engineering datasets often have a low mean train similarity and include labels that are not present in the training labels.

Additionally, Figure~\ref{fig:dataset_corr} includes the similarity values within each dataset. The similarity is calculated using the maximum cosine similarity for each label to the rest. Selecting the maximum value from all labels results in the inner max similarity, and a high value indicates that at least two labels in the task have very close embeddings. It corresponds to a high class similarity within our taxonomy. Therefore, these classes are challenging to differentiate, even without considering the image features (e.g., classes in MHP v1, Corrosion CS, and CUB-200). 

\section{Models}\label{sec:models}

We provide an overview of the tested zero-shot semantic segmentation models in Table~\ref{tab:model} including their modules and training datasets. We only include the datasets used for training the segmentation model and not the pre-training datasets of a utilized FM.
We want to point out that the public versions of X-Decoder and OpenSeeD are using different FMs than the larger, non-public versions. 

We utilize Grounded-SAM based on a re-implementation inspired by the demo code in~\cite{GroundedSAM2023}.
To our knowledge, other implementations of Grounded-SAM are limited to demo scripts and do not apply semantic segmentation. The model combines bounding box predictions from Grounding DINO~\cite{GroundingDINO2023} with instance segmentations from SAM~\cite{SegmentAnything}. Grounding DINO is an open-vocabulary object detection model. The model predicts bounding boxes for all class labels in the label set. The labels also include the background class, as we noticed better results in prior experiments compared to discarding the background class. Next, SAM predicts one instance segmentation mask for each bounding box, and the pixel-wise confidence values are scaled by the confidence score of the bounding box. The instance masks of each class are combined by the maximum confidence value of each pixel, resulting in semantic masks. Negative values represent background predictions. Therefore, pixels with only negative values are predicted as background or no-object for datasets without a background class. 

We noticed that Grounding DINO has a limited capability to predict non-general classes. SAM can also be combined with other open-vocabulary object detection models to improve performance. We refer to the oracle bounding box results for an upper bound.

\begin{table}[tbh]
    \caption{Overview of the evaluated models.}\label{tab:model}

    \begin{center}
    \small
    \setlength\extrarowheight{0.2em}
    \addtolength{\tabcolsep}{-0.2em}
    \begin{tabularx}{\linewidth}{lccYY}
\toprule
Name & Versions & Year & Modules & Training datasets \\
\midrule
ZSSeg \cite{ZSSeg2021} & Base & 2021 & CLIP ViT-B \& text encoder, Resnet 101 & COCO-Stuff\\
\hline
ZegFormer \cite{ZegFormer2022} & Base & 2022 & CLIP ViT-B \& text encoder, Resnet 101 & COCO-Stuff-156\\
\hline
OVSeg \cite{OVSeg2022} & Large & 2022 & CLIP ViT-L \& text encoder, Swin-B & COCO-Stuff, COCO Caption\\
\hline
X-Decoder \cite{XDecoder2022} & Tiny & 2023 & Focal-T/L, UniCL text encoder, ViT-decoder & COCO2017,  4M corpora\\
\hline
OpenSeeD \cite{OpenSeeD2023} & Tiny & 2023 & Swin-T, UniCL text encoder, ViT-decoder & COCO2017,  Objects365\\
\hline
SAN \cite{SAN2023} & B/L & 2023 & CLIP ViT-B/L \& text encoder, ViT-adapter & COCO-Stuff\\
\hline
CAT-Seg \cite{CAT-Seg2023} & B/L/H & 2023 & CLIP ViT-B/L/H \& text encoder, Swin-B, ViT-decoder & COCO-Stuff\\
\hline
Grounded-SAM \cite{GroundedSAM2023} & B/L/H & 2023 & SAM ViT-B/L/H, DINO Swin-B, BERT-B & SA-1B, COCO, O365, GoldG, Cap4M, OpenImage, ODinW-35, RefCOCO \\
\bottomrule
\end{tabularx}        
    \end{center}
\end{table}

\section{Additional results}\label{sec:experiments}

We provide further experiments and detailed dataset-wise results in this section. Specifically, we analyze classes of interest, the similarity between evaluation and training classes, and the influence of the segment size.

\subsection{Classes of interest}

\begin{wrapfigure}{r}{0.5\textwidth}
    \centering
    \includegraphics[width=0.45\textwidth]{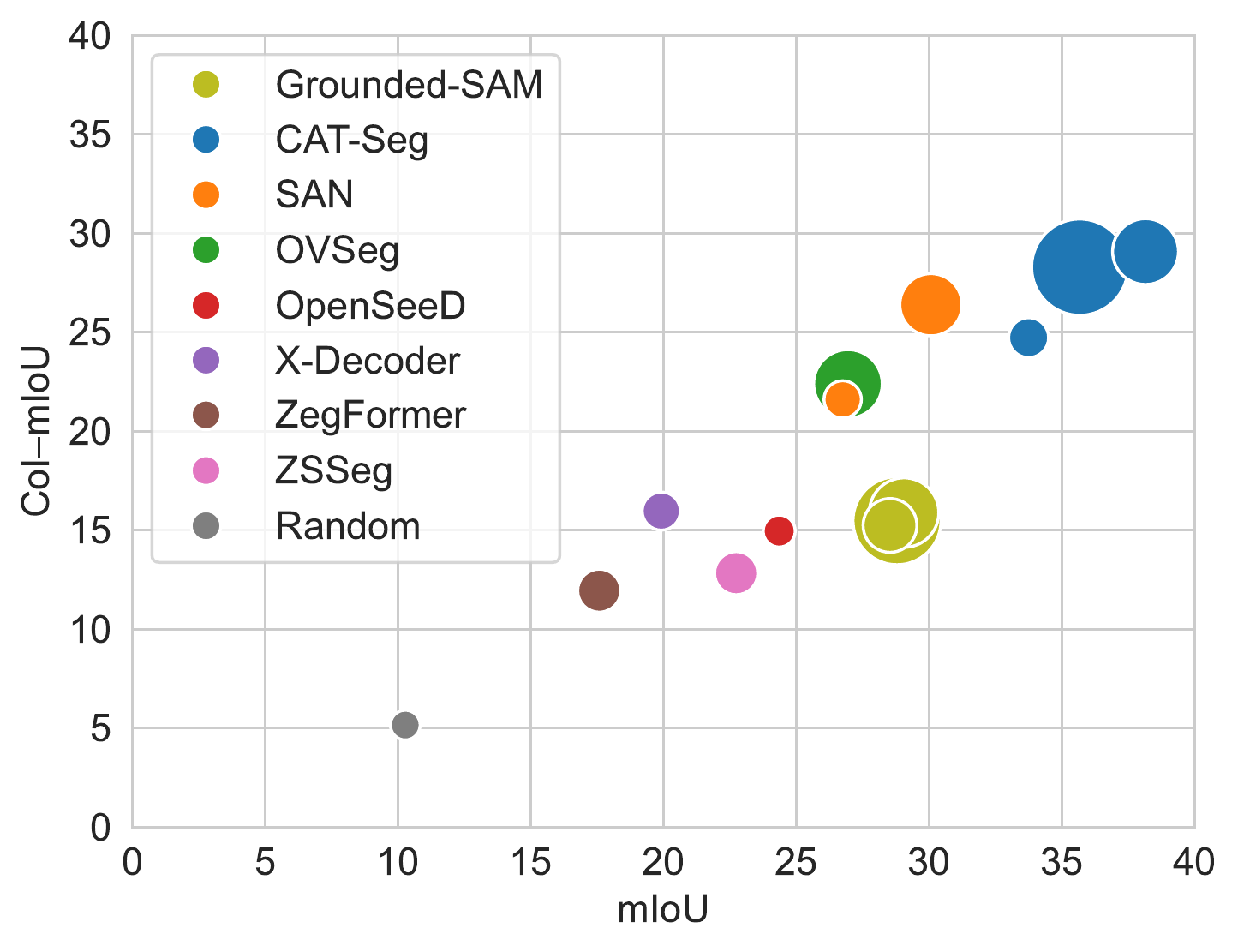}
    
  \caption{mIoU for Class(es) of Interest (CoI) in comparison to the mIoU of all classes. The size represents the parameter count of the models.}\label{fig:coi}
\end{wrapfigure}

For several datasets in the benchmark, a subset of the annotated classes is particularly relevant. We refer to this subset as Class(es) of Interest (CoI). E.g., binary segmentation tasks typically include a CoI (like pixels depicting a flood event) and a background class. In many cases, the model performance varies between CoI and background. To better understand the actual performance for these classes, we report the mIoU on the CoI subset (CoI-mIoU). With $CoI \subseteq \mathcal{C}$, and $IoU_i$ being the intersection over union for class $i$, we calculate the metric
$\textit{CoI-mIoU} = \frac{\sum_{i \in CoI}^{} IoU_i} {\mid CoI \mid}$. 
In binary segmentation tasks, this is similar to the $IoU_{pos}$ of the positive class~\cite{BenchmarkSAmedical}.

\begin{figure}[tbh]
    \centering  
    \subfloat[mIoU]{\includegraphics[width=0.48\linewidth]{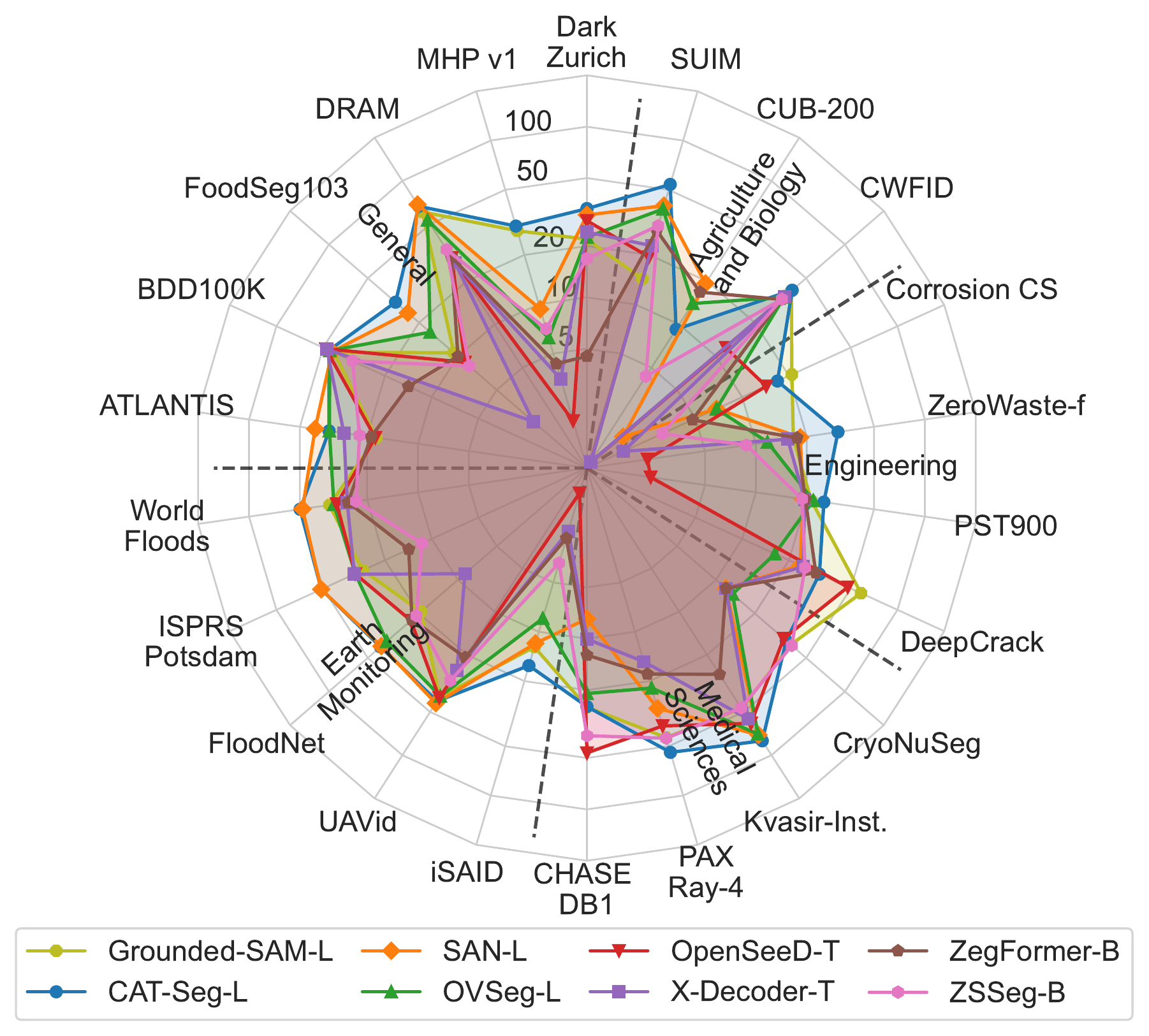}}
    \hfill
    \subfloat[CoI-mIoU]{\includegraphics[width=0.48\linewidth]{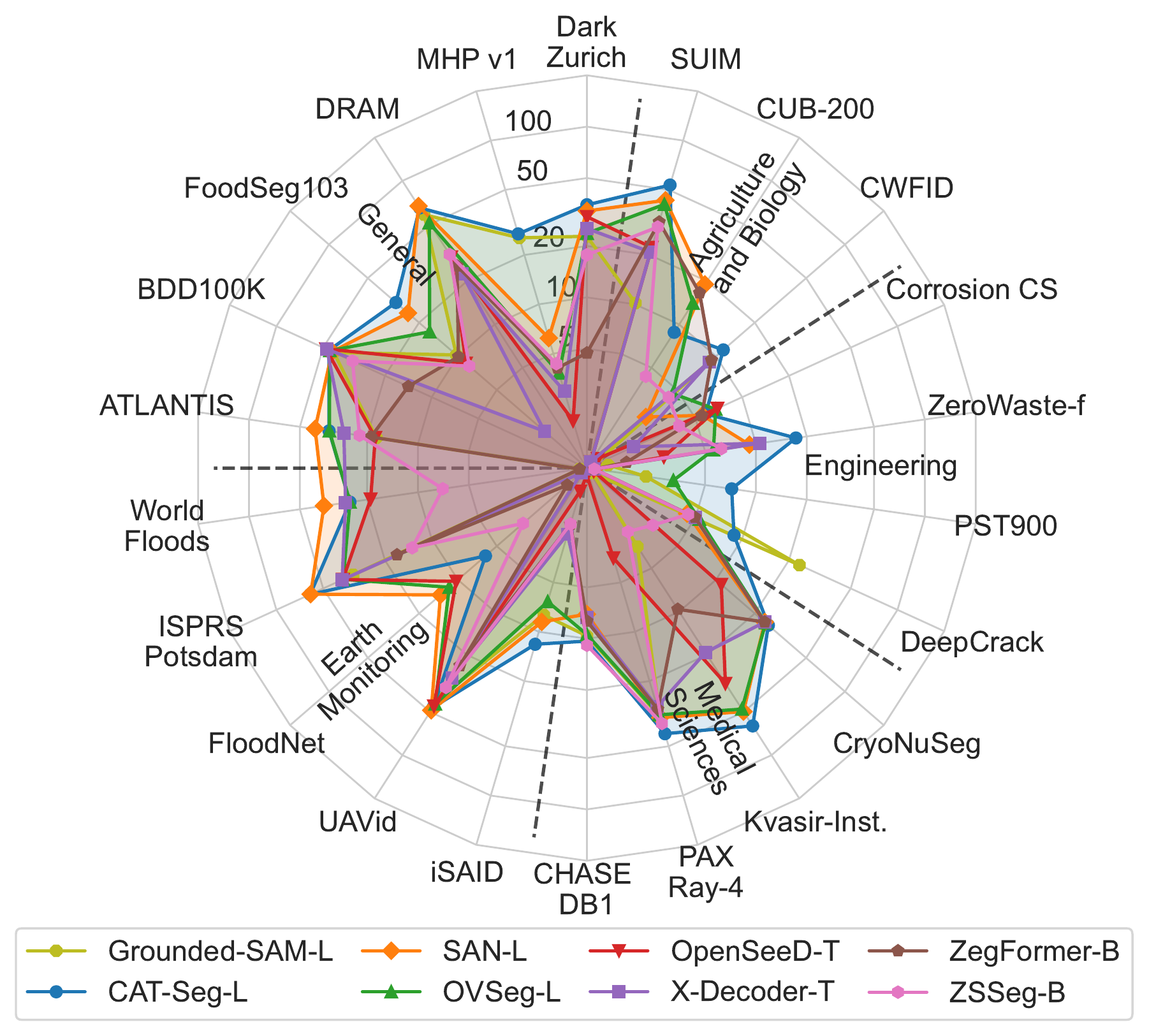}}
    
    \caption{mIoU (a) and CoI-mIoU (b) results for all model architectures on a log scale.}\label{fig:radar_coi}
\end{figure}

Figure \ref{fig:coi} visualizes the CoI-mIoU compared to the mIoU of all predicted classes. On average, none of the models is able to segment classes of interest as well as all classes. Models like ZSSeg, OpenSeeD, and Grounded-SAM have a particularly strong bias toward misclassifying CoI than the other models. Also, CAT-Seg tends to misclassify classes of interest. For example, SAN-L has on average only a 3.18pp lower CoI-mIoU than the best-performing model CAT-Seg-L while the mIoU difference is 8.08pp. 
The differences between the mIoU and the CoI-mIoU vary between the datasets and domains. Figure~\ref{fig:radar_coi} visualizes the mIoU and CoI-mIoU for all datasets and model architectures. The differences between both class sets are most evident for medical, engineering, and biological datasets (except for the datasets Kvasir-Instrument and SUIM). The CoI seem to be challenging for all models. These classes often include characteristics such as small segments, a high class similarity, or domain-specific labels. 
Furthermore, several model architectures tend to predict no or very few pixels as CoI, resulting in very low or zero scores. These architectures include X-Decoder, OpenSeeD, and Grounded-SAM. The models do not make use of CLIP, which may limit their capability to generalize to the domain-specific classes.

\subsection{Similarity to training classes}

\begin{figure}[tbh]
    \centering        
  
    \begin{minipage}[t]{.48\textwidth} 
    \centering
    \includegraphics[width=0.9\textwidth]{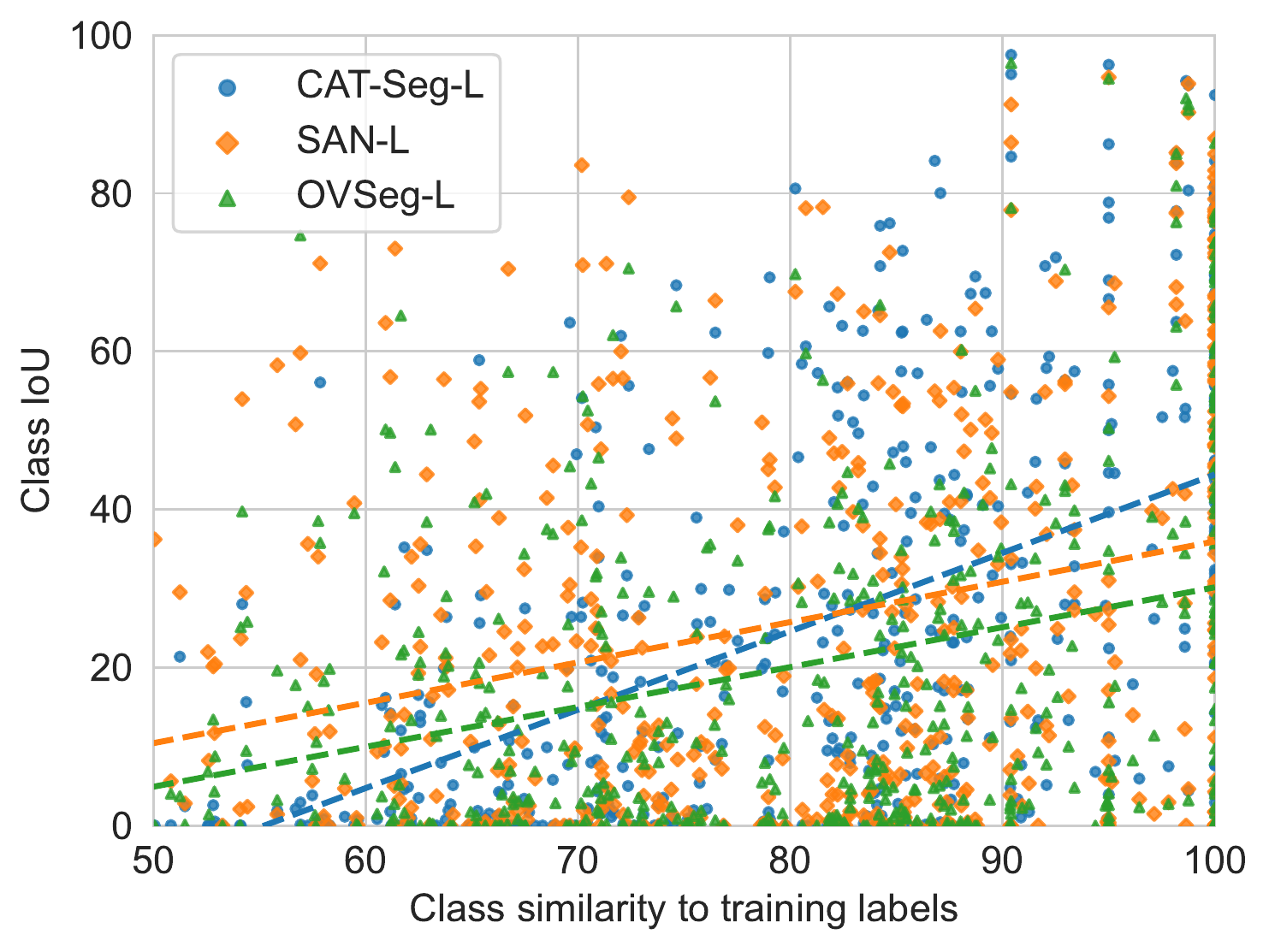}
    
    \caption{Class IoU in comparison to the class similarity with the labels in COCO-Stuff, represented by the maximum cosine similarity.}\label{fig:train_similarity}
    \end{minipage}
    \hspace*{\fill}     
    \begin{minipage}[t]{.48\textwidth}  
    \centering
     \includegraphics[width=0.9\textwidth]{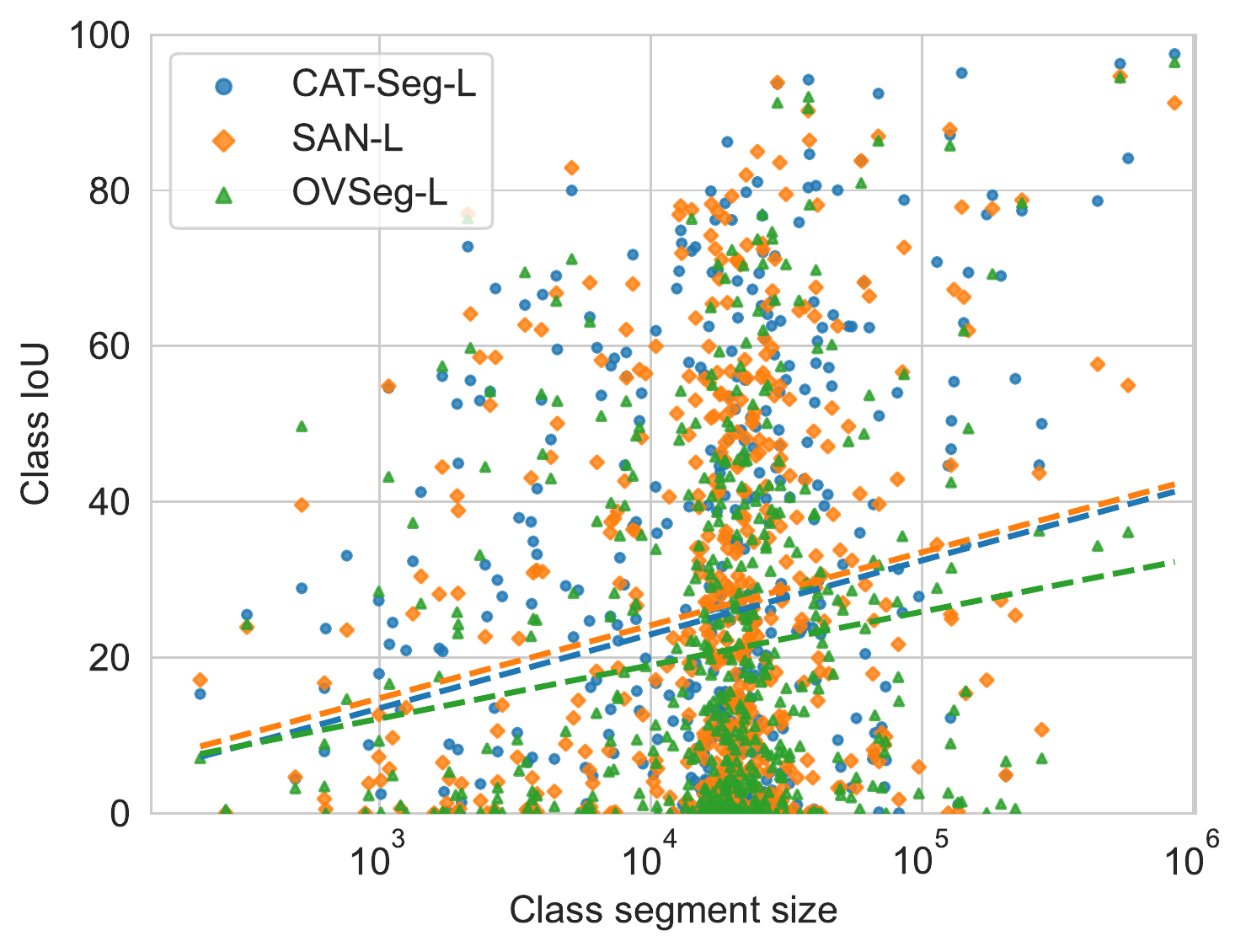}

    \caption{Class IoU in comparison to the class segment size on a log scale. The segment size is the class-wise average pixel count of a segment.}\label{fig:segment_size}
    \end{minipage}
\end{figure}

The generalized zero-shot transfer setting does allow an overlap between the training labels and the evaluation labels. We analyze this overlap and the influences on the model performance by calculating the embedding similarity of each label to the training labels in COCO-Stuff. A high similarity corresponds to the concept being present in the training dataset. Figure \ref{fig:train_similarity} presents the correlation between the similarity and the class-wise IoU for the three large models which are trained on COCO-Stuff. The results indicate a positive correlation between the similarity of the training labels and the performance. 
We also observe a comparable correlation for all other model architectures (except ZegFormer) which are partly trained on more diverse datasets.
The similarity of the training labels for the segmentation modules is not the only explanation. The correlation could be influenced by the open-vocabulary capabilities of the underlying FM. CLIP's understanding of common concepts, such as the training classes, is better than the understanding of domain-specific concepts~\cite{CLIP2021}. 

\subsection{Segment size}


Our benchmark includes multiple datasets with small segments, like WSI images with nuclei in cells or cars in satellite images. However, many models cannot correctly segment these small objects. We compare the average class segment size with the class IoU in Figure \ref{fig:segment_size}. The analysis considers all connected segments over 10 pixels to filter out potential annotation inaccuracies. 
Overall, all three large models have a positive correlation between segment size and mIoU––which also applies to other models. Therefore, the models have on average a lower performance on classes with small segments. 
We want to point out that the 200 CUB-200 classes are mostly correctly segmented but wrongly classified due to the challenging species labels. The correlation is higher without considering the CUB-200 classes.
Visual inspection leads to a second insight: Some models, e.g., CAT-Seg-L and SAN-L, are able to locate small objects but fail to correctly segment the boundaries. Therefore, nearby instances, like cars in satellite images, are often included in one segment.

\subsection{In-domain datasets}

We present the results of the five commonly used in-domain evaluation datasets in Table \ref{tab:results_common}. Some values differ from the officially reported performance, mostly within $\pm$1\%, which may be due to repeated runs~\cite{SAN2023}. It is worth noting that we could not reproduce the results from CAT-Seg on Pascal Context-459 and report a 4.2\% lower mIoU~\cite{CAT-Seg2023}. The results for Pascal VOC differ from the reported values in~\cite{CAT-Seg2023, SAN2023, OVSeg2022} because of a different evaluation setting. We included a 21st \textit{background} class and did not ignore the background pixels during evaluation. We find it misleading to ignore wrong predictions in the background, even if some objects are potentially not annotated. Other works assign the Pascal Context-59 labels that are not in PASCAL VOC to the background class~\cite{OpenSeg2021, CAT-Seg2023}. This may lead to better results than using the uniform label \textit{background}. 

Grounded-SAM has a very strong performance on Pascal VOC and nearly matches the fully-supervised result. However, the predictions become very noisy with an increasing number of classes, resulting in low mIoU scores. The CAT-Seg and SAN architectures produce the best results for the ADE20K and Pascal Context datasets.

\begin{table}[tbh]
\caption{mIoU results for all evaluated models on commonly used in-domain evaluation datasets.}\label{tab:results_common}

\begin{center}
\setlength\extrarowheight{0.1em}
\addtolength{\tabcolsep}{-0.4em}
\small
\begin{minipage}{\textwidth}
\begin{tabularx}{\linewidth}{lYYYYYY}
\toprule
 Model & ADE20K-150 & ADE20K-847 & Pascal Context-59 & Pascal Context-459 & Pascal VOC & Mean \\
\midrule
\textit{Random (LB)} & \textit{\phantom{0}0.16} & \textit{\phantom{0}0.02} & \textit{\phantom{0}0.6\phantom{0}} & \textit{\phantom{0}0.03} & \textit{\phantom{0}1.15} & \textit{\phantom{0}0.39} \\
\textit{Best supervised (UB)}
\footnote{The supervised models are InternImage-H~\cite{sotaADE20K} (ADE20K-150 and Pascal Context-59), MaskFormer~\cite{MaskFormer2021} (ADE20K-847), and  DeepLabv3+ (Xception-JFT)~\cite{sotaPascalVOC} (Pascal VOC). Pascal Context-459 is rarely used in supervised settings and has, to our knowledge, not been evaluated with recent models.}
& \textit{62.9\phantom{0}} & \textit{17.4\phantom{0}} & \textit{70.3\phantom{0}} & \textit{-}& \textit{84.56} & \textit{-} \\
\hline
\hline
ZSSeg-B & 19.85 & \phantom{0}4.91 & 47.5\phantom{0} & \phantom{0}8.81 & 42.27 & 24.67 \\
ZegFormer-B & 11.79 & \phantom{0}4.16 & 28.85 & \phantom{0}4.61 & 43.88 & 18.66 \\
X-Decoder-T & 25.13 & \phantom{0}6.37 & 54.19 & \phantom{0}9.72 & 38.13 & 26.71 \\
SAN-B & \textbf{27.56} & \textbf{10.22} & 54.07 & \underline{12.42} & 44.21 & \underline{29.7}\phantom{0} \\
OpenSeeD-T & 23.85 & \phantom{0}6.08 & \underline{56.79} & 12.19 & 39.17 & 27.61 \\
CAT-Seg-B & \underline{27.52} & \phantom{0}\underline{8.99} & \textbf{57.5}\phantom{0} & \textbf{13.47} & \underline{60.45} & \textbf{33.59} \\
Grounded-SAM-B & 14.75 & \phantom{0}2.58 & 41.65 & 10.05 & \textbf{77.19} & 29.25 \\
\hline
OVSeg-L & 29.58 & \phantom{0}9.11 & 55.32 & 12.07 & 40.82 & 29.38 \\
SAN-L & \underline{31.93} & \underline{12.92} & 57.53 & \textbf{16.31} & 50.16 & \underline{33.77} \\
CAT-Seg-L & 31.14 & 11.39 & \textbf{61.97} & \underline{16.2}\phantom{0} & 63.97 & \textbf{36.93} \\
Grounded-SAM-L & 15.18 & \phantom{0}2.58 & 44.02 & 10.75 & \textbf{82.36} & 30.98 \\
CAT-Seg-H & \textbf{34.52} & \textbf{13.08} & \underline{61.2}\phantom{0} & 16.03 & 43.53 & 33.67 \\
Grounded-SAM-H & 15.36 & \phantom{0}2.62 & 43.95 & 10.88 & \underline{81.51} & 30.86 \\
\bottomrule
\end{tabularx}

\end{minipage}
\end{center}
\end{table}

\subsection{Dataset-wise results}

Table \ref{tab:results_all} presents the mIoU results for all datasets. The best-performing model varies between the datasets. CAT-Seg is overall the best-performing model architecture, while SAN, Grounded-SAM, and OpenSeeD are better in some specific use cases. Table \ref{tab:results_coi} presents the CoI-mIoU results for each dataset. As discussed above, models without using CLIP often predict background instead of domain-specific classes which leads to a very low or zero CoI-mIoU.

\begin{table}[tbh]
\caption{mIoU results for all datasets grouped by their domain.}\label{tab:results_all}

\begin{center}
\setlength\extrarowheight{0.1em}
\addtolength{\tabcolsep}{-0.6em}
\tiny
\begin{tabularx}{\linewidth}{lYYYYYY|YYYYY|YYYY|YYYY|YYY|Y}
\toprule
 & \multicolumn{6}{c|}{General} & \multicolumn{5}{c|}{Earth Monitoring} & \multicolumn{4}{c|}{Medical Sciences} & \multicolumn{4}{c|}{Engineering} & \multicolumn{3}{c|}{Agri. and Biology} & \\
 & \rotatebox[origin=l]{90}{BDD100K} & \rotatebox[origin=l]{90}{Dark Zurich} & \rotatebox[origin=l]{90}{MHP v1} & \rotatebox[origin=l]{90}{FoodSeg103} & \rotatebox[origin=l]{90}{ATLANTIS} & \rotatebox[origin=l]{90}{DRAM} & \rotatebox[origin=l]{90}{iSAID} & \rotatebox[origin=l]{90}{ISPRS Pots.} & \rotatebox[origin=l]{90}{WorldFloods} & \rotatebox[origin=l]{90}{FloodNet} & \rotatebox[origin=l]{90}{UAVid} & \rotatebox[origin=l]{90}{Kvasir-Inst.} & \rotatebox[origin=l]{90}{CHASE DB1} & \rotatebox[origin=l]{90}{CryoNuSeg} & \rotatebox[origin=l]{90}{PAXRay-4} & \rotatebox[origin=l]{90}{Corrosion CS} & \rotatebox[origin=l]{90}{DeepCrack} & \rotatebox[origin=l]{90}{PST900} & \rotatebox[origin=l]{90}{ZeroWaste-f} & \rotatebox[origin=l]{90}{SUIM} & \rotatebox[origin=l]{90}{CUB-200} & \rotatebox[origin=l]{90}{CWFID} & \rotatebox[origin=l]{90}{Mean} \\
\midrule
\textit{Random (LB)} & \phantom{0}\textit{1.48} & \phantom{0}\textit{1.31} & \phantom{0}\textit{1.27} & \phantom{0}\textit{0.23} & \phantom{0}\textit{0.56} & \phantom{0}\textit{2.16} & \phantom{0}\textit{0.56} & \phantom{0}\textit{8.02} & \textit{18.43} & \phantom{0}\textit{3.39} & \phantom{0}\textit{5.18} & \textit{27.99} & \textit{27.25} & \textit{31.25} & \textit{31.53} & \textit{9.3} & \textit{26.52} & \phantom{0}\textit{4.52} & \phantom{0}\textit{6.49} & \phantom{0}\textit{5.3}\phantom{0} & \phantom{0}\textit{0.06} & \textit{13.08} & \textit{10.27} \\
\textit{Best sup. (UB)} & \textit{44.8}\phantom{0} & \textit{63.9}\phantom{0} & \textit{50.0}\phantom{0} & \textit{45.1}\phantom{0} & \textit{42.22} & \textit{45.71} & \textit{65.3}\phantom{0} & \textit{87.56} & \textit{92.71} & \textit{82.22} & \textit{67.8}\phantom{0} & \textit{93.7}\phantom{0} & \textit{97.05} & \textit{73.45} & \textit{93.77} & \textit{49.92} & \textit{85.9}\phantom{0} & \textit{82.3}\phantom{0} & \textit{52.5}\phantom{0} & \textit{74.0}\phantom{0} & \textit{84.6}\phantom{0} & \textit{87.23} & \textit{70.99} \\
\hline
\hline
ZSSeg-B & 32.36 & 16.86 & \phantom{0}7.08 & \phantom{0}8.17 & 22.19 & 33.19 & \phantom{0}3.8\phantom{0} & 11.57 & 23.25 & 20.98 & 30.27 & 46.93 & \underline{37.0}\phantom{0} & \textbf{38.7} & \underline{44.66} & \phantom{0}3.06 & 25.39 & 18.76 & \phantom{0}8.78 & \underline{30.16} & \phantom{0}4.35 & 32.46 & 22.73 \\
ZegFormer-B & 14.14 & \phantom{0}4.52 & \phantom{0}4.33 & 10.01 & 18.98 & 29.45 & \phantom{0}2.68 & 14.04 & 25.93 & 22.74 & 20.84 & 27.39 & 12.47 & 11.94 & 18.09 & \phantom{0}4.78 & 29.77 & 19.63 & 17.52 & 28.28 & \underline{16.8}\phantom{0} & 32.26 & 17.57 \\
X-Decoder-T & \underline{47.29} & 24.16 & \phantom{0}3.54 & \phantom{0}2.61 & 27.51 & 26.95 & \phantom{0}2.43 & 31.47 & 26.23 & \phantom{0}8.83 & 25.65 & 55.77 & 10.16 & 11.94 & 15.23 & \phantom{0}1.72 & 24.65 & 19.44 & 15.44 & 24.75 & \phantom{0}0.51 & 29.25 & 19.8\phantom{0} \\
SAN-B & 37.4\phantom{0} & 24.35 & \phantom{0}8.87 & \underline{19.27} & \textbf{36.51} & 49.68 & \phantom{0}4.77 & \underline{37.56} & 31.75 & \textbf{37.44} & \textbf{41.65} & \underline{69.88} & 17.85 & 11.95 & 19.73 & \phantom{0}3.13 & \underline{50.27} & 19.67 & \textbf{21.27} & 22.64 & \textbf{16.91} & \phantom{0}5.67 & 26.74 \\
OpenSeeD-T & \textbf{47.95} & \textbf{28.13} & \phantom{0}2.06 & \phantom{0}9.0\phantom{0} & 18.55 & 29.23 & \phantom{0}1.45 & 31.07 & 30.11 & 23.14 & 39.78 & 59.69 & \textbf{46.68} & 33.76 & 37.64 & 13.38 & 47.84 & \phantom{0}2.5\phantom{0} & \phantom{0}2.28 & 19.45 & \phantom{0}0.13 & 11.47 & 24.33 \\
CAT-Seg-B & 44.58 & \underline{27.36} & \underline{20.79} & \textbf{21.54} & \underline{33.08} & \textbf{62.42} & \textbf{15.75} & \textbf{41.89} & \textbf{39.47} & \underline{35.12} & \underline{40.62} & \textbf{70.68} & 25.38 & 25.63 & \textbf{44.94} & \underline{13.76} & 49.14 & \underline{21.32} & \underline{20.83} & \textbf{39.1}\phantom{0} & \phantom{0}3.4\phantom{0} & \textbf{45.47} & \textbf{33.74} \\
Gr.-SAM-B & 41.58 & 20.91 & \textbf{29.38} & 10.48 & 17.33 & \underline{57.38} & \underline{12.22} & 26.68 & \underline{33.41} & 19.19 & 38.34 & 46.82 & 23.56 & \underline{38.06} & 41.07 & \textbf{20.88} & \textbf{59.02} & \textbf{21.39} & 16.74 & 14.13 & \phantom{0}0.43 & \underline{38.41} & \underline{28.52} \\
\hline
OVSeg-L & 45.28 & 22.53 & \phantom{0}6.24 & 16.43 & 33.44 & 53.33 & \phantom{0}8.28 & 31.03 & 31.48 & 35.59 & 38.8 & 71.13 & 20.95 & 13.45 & 22.06 & \phantom{0}6.82 & 16.22 & \underline{21.89} & 11.71 & 38.17 & 14.0\phantom{0} & 33.76 & 26.94 \\
SAN-L & 43.81 & \underline{30.39} & \phantom{0}9.34 & 24.46 & \textbf{40.66} & \textbf{68.44} & 11.77 & \textbf{51.45} & \underline{48.24} & 39.26 & \textbf{43.41} & \underline{72.18} & \phantom{0}7.64 & 11.94 & 29.33 & \phantom{0}6.83 & 23.65 & 19.01 & 18.32 & 40.01 & \underline{19.3}\phantom{0} & \phantom{0}1.91 & 30.06 \\
CAT-Seg-L & \underline{45.83} & \textbf{33.1}\phantom{0} & \textbf{30.03} & \textbf{30.47} & 33.6\phantom{0} & \underline{66.54} & \textbf{16.09} & \underline{51.42} & \textbf{49.86} & \underline{39.84} & \underline{42.02} & \textbf{79.4} & 24.99 & 35.06 & \textbf{54.5}\phantom{0} & 16.87 & 31.42 & \textbf{25.26} & \textbf{30.62} & \textbf{53.94} & \phantom{0}9.24 & \underline{39.0}\phantom{0} & \textbf{38.14} \\
Gr.-SAM-L & 42.69 & 21.92 & \underline{28.11} & 10.76 & 17.63 & 60.8\phantom{0} & 12.38 & 27.76 & 33.4\phantom{0} & 19.28 & 39.37 & 47.32 & \textbf{25.16} & \textbf{38.06} & \underline{44.22} & \textbf{20.88} & \textbf{58.21} & 21.23 & 16.67 & 14.3\phantom{0} & \phantom{0}0.43 & 38.47 & 29.05 \\
CAT-Seg-H & \textbf{48.34} & 29.72 & 23.53 & \underline{29.06} & \underline{40.43} & 56.78 & \phantom{0}9.04 & 49.37 & 47.92 & \textbf{40.98} & 41.36 & 70.7\phantom{0} & 13.37 & 12.82 & 41.72 & 12.17 & \underline{57.69} & 19.61 & \underline{26.71} & \underline{47.8}\phantom{0} & \textbf{19.49} & \textbf{45.99} & \underline{35.66} \\
Gr.-SAM-H & 42.95 & 22.09 & 28.05 & \phantom{0}9.97 & 17.68 & 60.86 & \underline{12.44} & 27.79 & 33.23 & 19.31 & 39.41 & 46.97 & \underline{25.13} & \textbf{38.06} & 43.64 & \textbf{20.88} & 53.74 & 21.34 & 16.68 & 14.3\phantom{0} & \phantom{0}0.43 & 38.29 & 28.78 \\
\bottomrule
\end{tabularx}

\end{center}
\end{table}

\begin{table}[tbh]
\caption{CoI-mIoU results for all datasets grouped by their domain.}\label{tab:results_coi}

\begin{center}
\setlength\extrarowheight{0.1em}
\addtolength{\tabcolsep}{-0.6em}
\tiny
\begin{tabularx}{\linewidth}{lYYYYYY|YYYYY|YYYY|YYYY|YYY|Y}
\toprule
 & \multicolumn{6}{c|}{General} & \multicolumn{5}{c|}{Earth Monitoring} & \multicolumn{4}{c|}{Medical Sciences} & \multicolumn{4}{c|}{Engineering} & \multicolumn{3}{c|}{Agri. and Biology} & \\
 & \rotatebox[origin=l]{90}{BDD100K} & \rotatebox[origin=l]{90}{Dark Zurich} & \rotatebox[origin=l]{90}{MHP v1} & \rotatebox[origin=l]{90}{FoodSeg103} & \rotatebox[origin=l]{90}{ATLANTIS} & \rotatebox[origin=l]{90}{DRAM} & \rotatebox[origin=l]{90}{iSAID} & \rotatebox[origin=l]{90}{ISPRS Pots.} & \rotatebox[origin=l]{90}{WorldFloods} & \rotatebox[origin=l]{90}{FloodNet} & \rotatebox[origin=l]{90}{UAVid} & \rotatebox[origin=l]{90}{Kvasir-Inst.} & \rotatebox[origin=l]{90}{CHASE DB1} & \rotatebox[origin=l]{90}{CryoNuSeg} & \rotatebox[origin=l]{90}{PAXRay-4} & \rotatebox[origin=l]{90}{Corrosion CS} & \rotatebox[origin=l]{90}{DeepCrack} & \rotatebox[origin=l]{90}{PST900} & \rotatebox[origin=l]{90}{ZeroWaste-f} & \rotatebox[origin=l]{90}{SUIM} & \rotatebox[origin=l]{90}{CUB-200} & \rotatebox[origin=l]{90}{CWFID} & \rotatebox[origin=l]{90}{Mean} \\
\midrule
\textit{Random (LB)} & \textit{\phantom{0}1.48} & \textit{\phantom{0}1.28} & \textit{\phantom{0}1.06} & \textit{\phantom{0}0.22} & \textit{\phantom{0}0.56} & \textit{\phantom{0}1.62} & \textit{\phantom{0}0.18} & \textit{\phantom{0}8.87} & \textit{15.35} & \textit{\phantom{0}1.83} & \textit{\phantom{0}4.84} & \textit{\phantom{0}8.38} & \textit{\phantom{0}6.22} & \textit{19.28} & \textit{21.58} & \textit{\phantom{0}4.46} & \textit{\phantom{0}4.15} & \textit{\phantom{0}0.67} & \textit{\phantom{0}3.33} & \textit{\phantom{0}4.53} & \textit{\phantom{0}0.06} & \textit{\phantom{0}3.38} & \textit{\phantom{0}5.15} \\
\hline
\hline
ZSSeg-B & 32.36 & 17.75 & \phantom{0}4.33 & \phantom{0}8.16 & 22.19 & 30.71 & \phantom{0}2.2 & 13.35 & \phantom{0}7.13 & \phantom{0}3.12 & 33.74 & \phantom{0}2.77 & \textbf{10.93} & \phantom{0}3.25 & \underline{36.3} & \phantom{0}3.92 & \phantom{0}4.49 & \phantom{0}0.93 & \phantom{0}6.24 & 29.63 & \phantom{0}4.35 & \phantom{0}4.29 & 12.83 \\
ZegFormer-B & 14.14 & \phantom{0}4.72 & \phantom{0}4.08 & \phantom{0}9.91 & 18.98 & 25.6\phantom{0} & \phantom{0}2.2 & 16.72 & \phantom{0}0.0\phantom{0} & \phantom{0}1.42 & 23.81 & \phantom{0}9.63 & \phantom{0}7.89 & 23.88 & 29.75 & \phantom{0}\underline{5.49} & \phantom{0}4.96 & \phantom{0}0.24 & \phantom{0}1.71 & \underline{31.8}\phantom{0} & \underline{16.6} & \phantom{0}\underline{9.24} & 11.94 \\
X-Decoder-T & \underline{47.29} & 25.3\phantom{0} & \phantom{0}2.98 & \phantom{0}2.13 & 27.51 & 22.55 & \phantom{0}2.54 & 37.71 & \textbf{26.84} & \phantom{0}0.77 & 28.95 & 19.25 & \phantom{0}7.54 & 23.88 & 28.73 & \phantom{0}2.0\phantom{0} & \phantom{0}4.98 & \phantom{0}0.0\phantom{0} & \textbf{10.52} & 22.28 & \phantom{0}0.07 & \phantom{0}7.96 & 15.99 \\
SAN-B & 37.4\phantom{0} & 25.63 & \phantom{0}6.32 & \underline{19.16} & \textbf{36.51} & 47.7\phantom{0} & \phantom{0}4.55 & \underline{45.0}\phantom{0} & \underline{20.01} & \textbf{14.41} & \textbf{46.08} & \underline{45.69} & \phantom{0}8.86 & \underline{23.89} & 30.18 & \phantom{0}3.48 & \phantom{0}6.5\phantom{0} & \phantom{0}1.35 & \phantom{0}7.0\phantom{0} & 25.52 & \textbf{16.82} & \phantom{0}3.17 & \underline{21.6}\phantom{0} \\
OpenSeeD-T & \textbf{47.95} & \textbf{29.7}\phantom{0} & \phantom{0}2.03 & \phantom{0}8.81 & 18.55 & 29.62 & \phantom{0}1.41 & 37.28 & 19.26 & \underline{10.32} & \underline{45.46} & 31.38 & \phantom{0}0.0\phantom{0} & \phantom{0}8.97 & \phantom{0}3.69 & \phantom{0}\textbf{5.8}\phantom{0} & \phantom{0}0.0\phantom{0} & \phantom{0}0.17 & \phantom{0}2.85 & 22.16 & \phantom{0}0.13 & \phantom{0}1.19 & 14.85 \\
CAT-Seg-B & 44.58 & \underline{28.8}\phantom{0} & \underline{17.05} & \textbf{21.28} & \underline{33.08} & \textbf{60.26} & \textbf{13.16} & \textbf{50.07} & \phantom{0}5.74 & \phantom{0}6.74 & 45.09 & \textbf{47.66} & \underline{10.35} & \textbf{25.98} & \textbf{39.78} & \phantom{0}5.12 & \underline{17.63} & \phantom{0}\underline{2.38} & \phantom{0}\underline{7.84} & \textbf{37.49} & \phantom{0}2.93 & \textbf{20.88} & \textbf{24.72} \\
Gr.-SAM-B & 41.58 & 21.75 & \textbf{26.7}\phantom{0} & 10.01 & 17.33 & \underline{54.66} & \phantom{0}\underline{7.73} & 30.7\phantom{0} & \phantom{0}0.0\phantom{0} & \phantom{0}0.0\phantom{0} & 39.42 & \phantom{0}2.71 & \phantom{0}9.71 & \phantom{0}0.0\phantom{0} & 26.52 & \phantom{0}0.0\phantom{0} & \textbf{23.72} & \phantom{0}\textbf{2.42} & \phantom{0}1.39 & \phantom{0}9.99 & \phantom{0}0.0\phantom{0} & \phantom{0}8.9\phantom{0} & 15.24 \\
\hline
OVSeg-L & 45.28 & 23.72 & \phantom{0}3.8\phantom{0} & 16.56 & 33.44 & 51.07 & \phantom{0}6.54 & 37.13 & 25.27 & \underline{11.67} & 44.02 & 47.77 & \phantom{0}9.46 & \underline{24.29} & 32.13 & \phantom{0}\textbf{6.75} & \phantom{0}5.29 & \phantom{0}\underline{3.25} & \phantom{0}5.61 & 40.75 & 14.06 & \phantom{0}4.64 & 22.39 \\
SAN-L & 43.81 & \underline{32.08} & \phantom{0}6.22 & 24.37 & \textbf{40.66} & \textbf{66.81} & \phantom{0}\underline{8.71} & \underline{60.17} & \textbf{36.03} & \textbf{13.65} & \textbf{48.67} & \underline{49.69} & \phantom{0}7.18 & 23.88 & 33.44 & \phantom{0}5.54 & \phantom{0}4.42 & \phantom{0}0.96 & \phantom{0}9.16 & 43.17 & \underline{19.0}\phantom{0} & \phantom{0}2.86 & 26.39 \\
CAT-Seg-L & \underline{45.83} & \textbf{34.84} & \textbf{26.91} & \textbf{30.26} & 33.6\phantom{0} & \underline{64.89} & \textbf{11.92} & \textbf{60.53} & 25.28 & \phantom{0}6.11 & \underline{46.32} & \textbf{62.54} & \textbf{10.33} & \textbf{25.49} & \textbf{41.91} & \phantom{0}\underline{5.82} & \phantom{0}8.85 & \phantom{0}\textbf{7.19} & \textbf{17.24} & \underline{53.47} & \phantom{0}8.82 & \underline{11.4}\phantom{0} & \textbf{29.07} \\
Gr.-SAM-L & 42.69 & 22.8\phantom{0} & \underline{25.44} & 10.28 & 17.63 & 58.18 & \phantom{0}7.89 & 32.0\phantom{0} & \phantom{0}0.0\phantom{0} & \phantom{0}0.0\phantom{0} & 40.35 & \phantom{0}3.52 & \phantom{0}\underline{9.63} & \phantom{0}0.0\phantom{0} & 32.92 & \phantom{0}0.0\phantom{0} & \textbf{23.39} & \phantom{0}2.24 & \phantom{0}1.34 & 10.18 & \phantom{0}0.0\phantom{0} & \phantom{0}8.99 & 15.89 \\
CAT-Seg-H & \textbf{48.34} & 31.29 & 20.61 & \underline{28.92} & \underline{40.43} & 55.02 & \phantom{0}8.31 & 58.91 & \underline{26.92} & 11.49 & 45.88 & 46.67 & \phantom{0}8.04 & 23.74 & \underline{33.47} & \phantom{0}4.09 & \underline{19.4} & \phantom{0}1.27 & \underline{14.08} & \textbf{53.92} & \textbf{19.42} & \textbf{22.02} & \underline{28.28} \\
Gr.-SAM-H & 42.95 & 22.97 & 25.4\phantom{0} & \phantom{0}9.49 & 17.68 & 58.25 & \phantom{0}7.85 & 32.02 & \phantom{0}0.0\phantom{0} & \phantom{0}0.0\phantom{0} & 40.44 & \phantom{0}2.86 & \phantom{0}\underline{9.63} & \phantom{0}0.0\phantom{0} & 31.75 & \phantom{0}0.0\phantom{0} & 16.47 & \phantom{0}2.35 & \phantom{0}1.34 & 10.18 & \phantom{0}0.0\phantom{0} & \phantom{0}8.81 & 15.47 \\
\bottomrule
\end{tabularx}

\end{center}
\end{table}

\section{Qualitative examples}\label{sec:predictions}

\begin{figure}[h]
    \centering
    \footnotesize    \addtolength{\tabcolsep}{-0.5em}
    \begin{tabular}{lcccccc}
 & Image & Ground Truth & CAT-Seg-L & SAN-L & OVSeg-L & Gr.-SAM-L \\
\rotatebox[origin=l]{90}{BDD.} & \includegraphics[width=.15\textwidth]{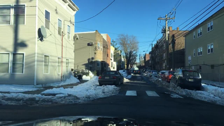} & \includegraphics[width=.15\textwidth]{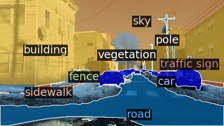} & \includegraphics[width=.15\textwidth]{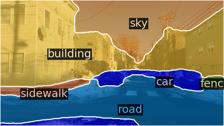} & \includegraphics[width=.15\textwidth]{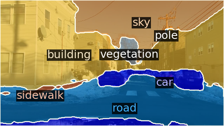} & \includegraphics[width=.15\textwidth]{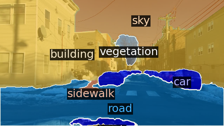} & \includegraphics[width=.15\textwidth]{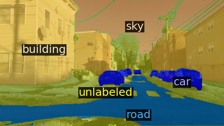} \\
\rotatebox[origin=l]{90}{Dark Z.} & \includegraphics[width=.15\textwidth]{predictions/CAT-Seg_dark_zurich_sem_seg_val_1_image.png} & \includegraphics[width=.15\textwidth]{predictions/CAT-Seg_dark_zurich_sem_seg_val_1_gt.png} & \includegraphics[width=.15\textwidth]{predictions/CAT-Seg_dark_zurich_sem_seg_val_1_prediction.png} & \includegraphics[width=.15\textwidth]{predictions/SAN_dark_zurich_sem_seg_val_1_prediction.png} & \includegraphics[width=.15\textwidth]{predictions/ovseg_dark_zurich_sem_seg_val_1_prediction.png} & \includegraphics[width=.15\textwidth]{predictions/Grounded-SAM_large_dark_zurich_sem_seg_val_1_prediction.png} \\
\rotatebox[origin=l]{90}{MHP v1} & \includegraphics[width=.15\textwidth]{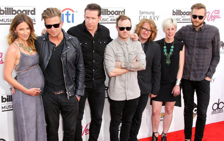} & \includegraphics[width=.15\textwidth]{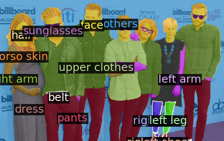} & \includegraphics[width=.15\textwidth]{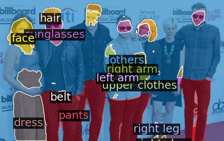} & \includegraphics[width=.15\textwidth]{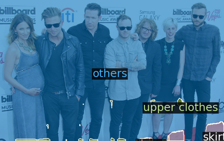} & \includegraphics[width=.15\textwidth]{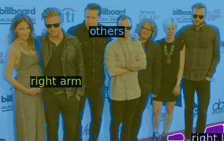} & \includegraphics[width=.15\textwidth]{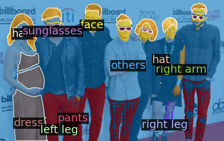} \\
\rotatebox[origin=l]{90}{FoodSeg103} & \includegraphics[width=.15\textwidth]{predictions/CAT-Seg_foodseg103_sem_seg_test_1_image.png} & \includegraphics[width=.15\textwidth]{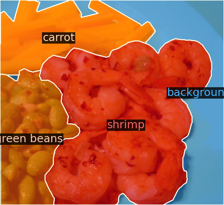} & \includegraphics[width=.15\textwidth]{predictions/CAT-Seg_foodseg103_sem_seg_test_1_prediction.png} & \includegraphics[width=.15\textwidth]{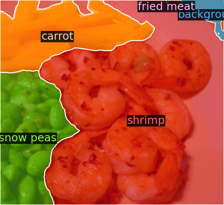} & \includegraphics[width=.15\textwidth]{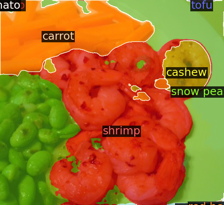} & \includegraphics[width=.15\textwidth]{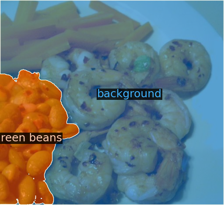} \\
\rotatebox[origin=l]{90}{ATLAN.} & \includegraphics[width=.15\textwidth]{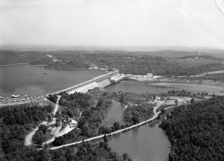} & \includegraphics[width=.15\textwidth]{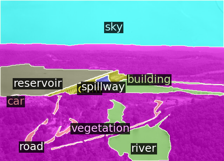} & \includegraphics[width=.15\textwidth]{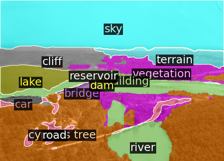} & \includegraphics[width=.15\textwidth]{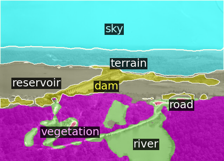} & \includegraphics[width=.15\textwidth]{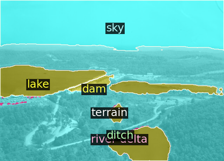} & \includegraphics[width=.15\textwidth]{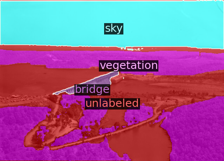} \\
\rotatebox[origin=l]{90}{DRAM} & \includegraphics[width=.15\textwidth]{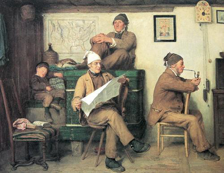} & \includegraphics[width=.15\textwidth]{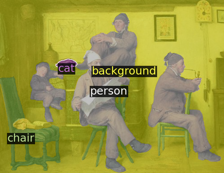} & \includegraphics[width=.15\textwidth]{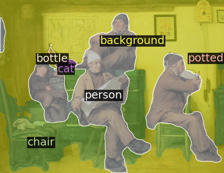} & \includegraphics[width=.15\textwidth]{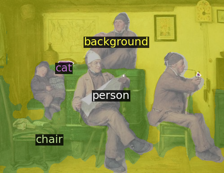} & \includegraphics[width=.15\textwidth]{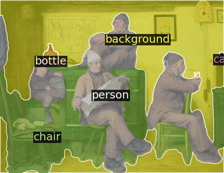} & \includegraphics[width=.15\textwidth]{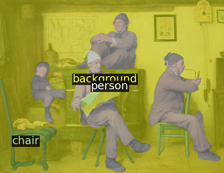} \\
\rotatebox[origin=l]{90}{iSAID} & \includegraphics[width=.15\textwidth]{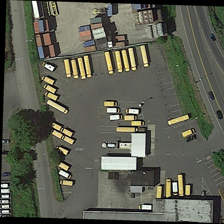} & \includegraphics[width=.15\textwidth]{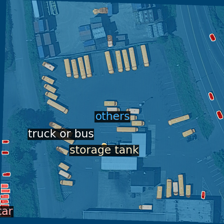} & \includegraphics[width=.15\textwidth]{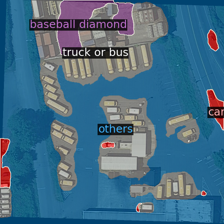} & \includegraphics[width=.15\textwidth]{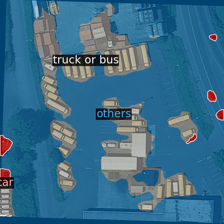} & \includegraphics[width=.15\textwidth]{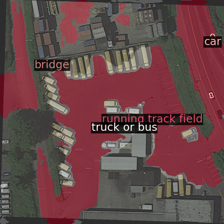} & \includegraphics[width=.15\textwidth]{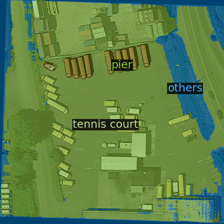} \\
\rotatebox[origin=l]{90}{ISPRS Pots.} & \includegraphics[width=.15\textwidth]{predictions/CAT-Seg_isprs_potsdam_sem_seg_test_irrg_6_image.png} & \includegraphics[width=.15\textwidth]{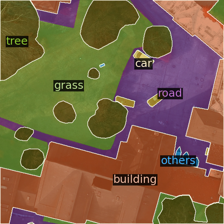} & \includegraphics[width=.15\textwidth]{predictions/CAT-Seg_isprs_potsdam_sem_seg_test_irrg_6_prediction.png} & \includegraphics[width=.15\textwidth]{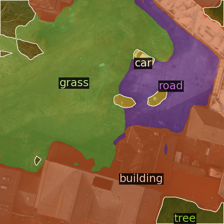} & \includegraphics[width=.15\textwidth]{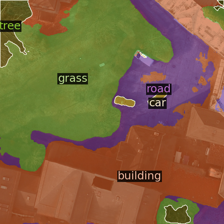} & \includegraphics[width=.15\textwidth]{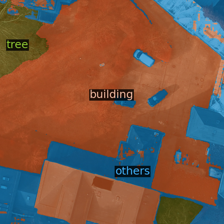} \\
\rotatebox[origin=l]{90}{WorldFloods} & \includegraphics[width=.15\textwidth]{predictions/CAT-Seg_worldfloods_sem_seg_test_irrg_1_image.png} & \includegraphics[width=.15\textwidth]{predictions/CAT-Seg_worldfloods_sem_seg_test_irrg_1_gt.png} & \includegraphics[width=.15\textwidth]{predictions/CAT-Seg_worldfloods_sem_seg_test_irrg_1_prediction.png} & \includegraphics[width=.15\textwidth]{predictions/SAN_worldfloods_sem_seg_test_irrg_1_prediction.png} & \includegraphics[width=.15\textwidth]{predictions/ovseg_worldfloods_sem_seg_test_irrg_1_prediction.png} & \includegraphics[width=.15\textwidth]{predictions/Grounded-SAM_large_worldfloods_sem_seg_test_irrg_1_prediction.png} \\
\rotatebox[origin=l]{90}{FloodNet} & \includegraphics[width=.15\textwidth]{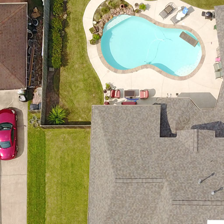} & \includegraphics[width=.15\textwidth]{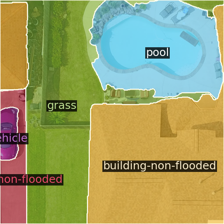} & \includegraphics[width=.15\textwidth]{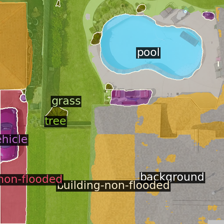} & \includegraphics[width=.15\textwidth]{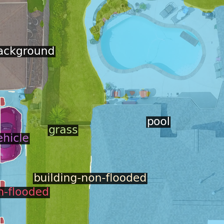} & \includegraphics[width=.15\textwidth]{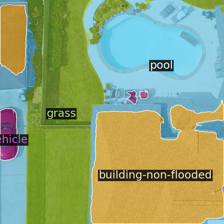} & \includegraphics[width=.15\textwidth]{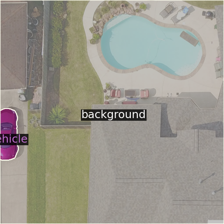} \\
\rotatebox[origin=l]{90}{UAVid} & \includegraphics[width=.15\textwidth]{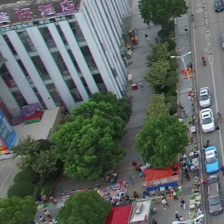} & \includegraphics[width=.15\textwidth]{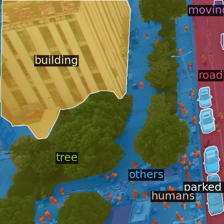} & \includegraphics[width=.15\textwidth]{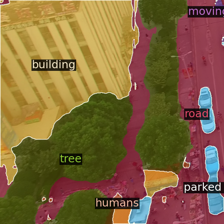} & \includegraphics[width=.15\textwidth]{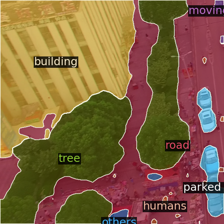} & \includegraphics[width=.15\textwidth]{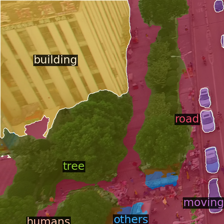} & \includegraphics[width=.15\textwidth]{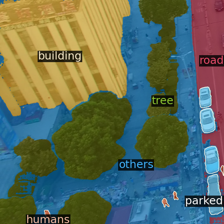} \\
\end{tabular}

    \caption{Predictions for datasets of the domains general and earth monitoring.}
    \label{fig:predictions1}
\end{figure}

\begin{figure}[h]
    \centering
    \footnotesize
    \addtolength{\tabcolsep}{-0.5em}    \begin{tabular}{lcccccc}
 & Image & Ground Truth & CAT-Seg-L & SAN-L & OVSeg-L & Gr.-SAM-L \\
\rotatebox[origin=l]{90}{Kvasir-Inst.} & \includegraphics[width=.15\textwidth]{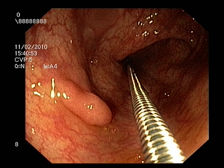} & \includegraphics[width=.15\textwidth]{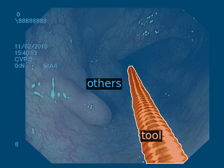} & \includegraphics[width=.15\textwidth]{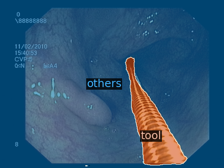} & \includegraphics[width=.15\textwidth]{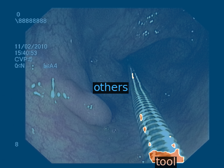} & \includegraphics[width=.15\textwidth]{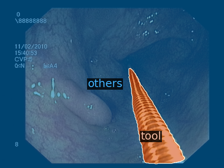} & \includegraphics[width=.15\textwidth]{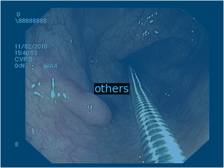} \\
\rotatebox[origin=l]{90}{CHASE DB1} & \includegraphics[width=.15\textwidth]{predictions/CAT-Seg_chase_db1_sem_seg_test_3_image.png} & \includegraphics[width=.15\textwidth]{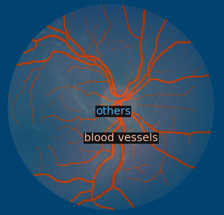} & \includegraphics[width=.15\textwidth]{predictions/CAT-Seg_chase_db1_sem_seg_test_3_prediction.png} & \includegraphics[width=.15\textwidth]{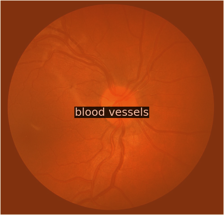} & \includegraphics[width=.15\textwidth]{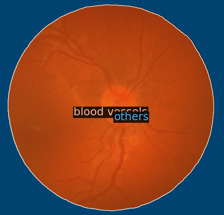} & \includegraphics[width=.15\textwidth]{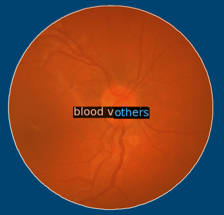} \\
\rotatebox[origin=l]{90}{CryoNuSeg} & \includegraphics[width=.15\textwidth]{predictions/CAT-Seg_cryonuseg_sem_seg_test_1_image.png} & \includegraphics[width=.15\textwidth]{predictions/CAT-Seg_cryonuseg_sem_seg_test_1_gt.png} & \includegraphics[width=.15\textwidth]{predictions/CAT-Seg_cryonuseg_sem_seg_test_1_prediction.png} & \includegraphics[width=.15\textwidth]{predictions/SAN_cryonuseg_sem_seg_test_1_prediction.png} & \includegraphics[width=.15\textwidth]{predictions/ovseg_cryonuseg_sem_seg_test_1_prediction.png} & \includegraphics[width=.15\textwidth]{predictions/Grounded-SAM_large_cryonuseg_sem_seg_test_1_prediction.png} \\
\rotatebox[origin=l]{90}{PAXRay-Lungs} & \includegraphics[width=.15\textwidth]{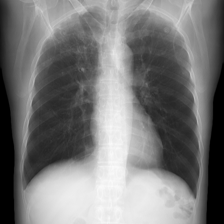} & \includegraphics[width=.15\textwidth]{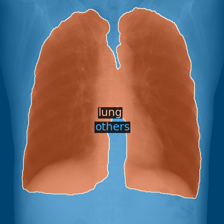} & \includegraphics[width=.15\textwidth]{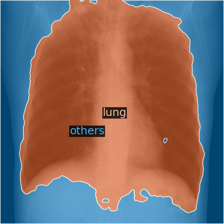} & \includegraphics[width=.15\textwidth]{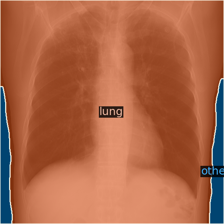} & \includegraphics[width=.15\textwidth]{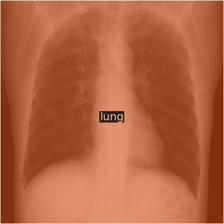} & \includegraphics[width=.15\textwidth]{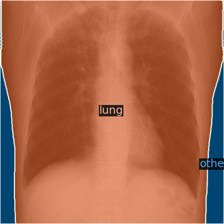} \\
\rotatebox[origin=l]{90}{Corr. CS} & \includegraphics[width=.15\textwidth]{predictions/CAT-Seg_corrosion_cs_sem_seg_test_0_image.png} & \includegraphics[width=.15\textwidth]{predictions/CAT-Seg_corrosion_cs_sem_seg_test_0_gt.png} & \includegraphics[width=.15\textwidth]{predictions/CAT-Seg_corrosion_cs_sem_seg_test_0_prediction.png} & \includegraphics[width=.15\textwidth]{predictions/SAN_corrosion_cs_sem_seg_test_0_prediction.png} & \includegraphics[width=.15\textwidth]{predictions/ovseg_corrosion_cs_sem_seg_test_0_prediction.png} & \includegraphics[width=.15\textwidth]{predictions/Grounded-SAM_large_corrosion_cs_sem_seg_test_0_prediction.png} \\
\rotatebox[origin=l]{90}{DeepCrack} & \includegraphics[width=.15\textwidth]{predictions/CAT-Seg_deepcrack_sem_seg_test_6_image.png} & \includegraphics[width=.15\textwidth]{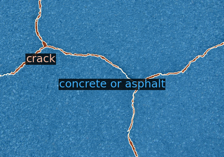} & \includegraphics[width=.15\textwidth]{predictions/CAT-Seg_deepcrack_sem_seg_test_6_prediction.png} & \includegraphics[width=.15\textwidth]{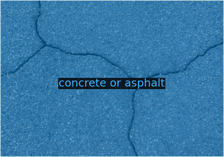} & \includegraphics[width=.15\textwidth]{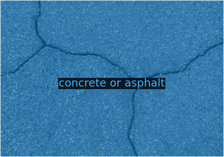} & \includegraphics[width=.15\textwidth]{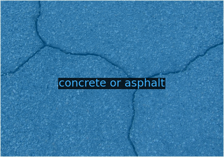} \\
\rotatebox[origin=l]{90}{PST900} & \includegraphics[width=.15\textwidth]{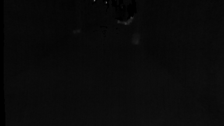} & \includegraphics[width=.15\textwidth]{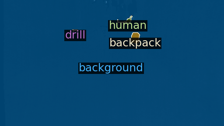} & \includegraphics[width=.15\textwidth]{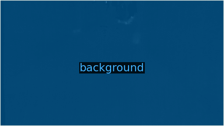} & \includegraphics[width=.15\textwidth]{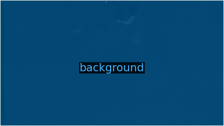} & \includegraphics[width=.15\textwidth]{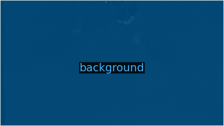} & \includegraphics[width=.15\textwidth]{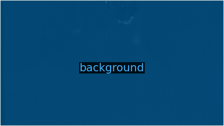} \\
\rotatebox[origin=l]{90}{ZeroW.} & \includegraphics[width=.15\textwidth]{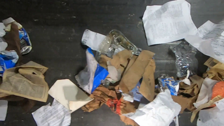} & \includegraphics[width=.15\textwidth]{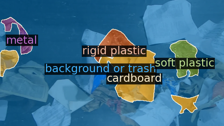} & \includegraphics[width=.15\textwidth]{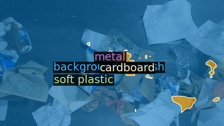} & \includegraphics[width=.15\textwidth]{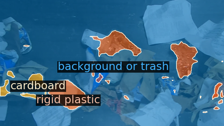} & \includegraphics[width=.15\textwidth]{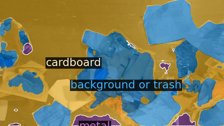} & \includegraphics[width=.15\textwidth]{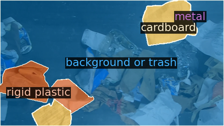} \\
\rotatebox[origin=l]{90}{SUIM} & \includegraphics[width=.15\textwidth]{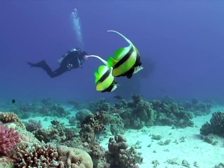} & \includegraphics[width=.15\textwidth]{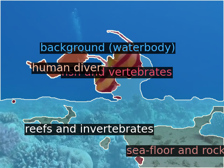} & \includegraphics[width=.15\textwidth]{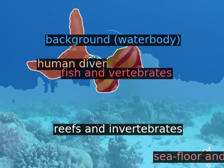} & \includegraphics[width=.15\textwidth]{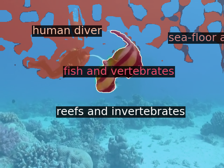} & \includegraphics[width=.15\textwidth]{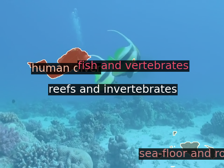} & \includegraphics[width=.15\textwidth]{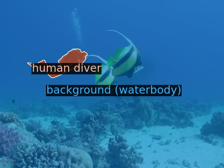} \\
\rotatebox[origin=l]{90}{CUB-200} & \includegraphics[width=.15\textwidth]{predictions/CAT-Seg_cub_200_sem_seg_test_2_image.png} & \includegraphics[width=.15\textwidth]{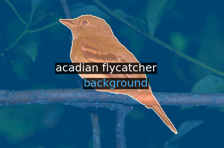} & \includegraphics[width=.15\textwidth]{predictions/CAT-Seg_cub_200_sem_seg_test_2_prediction.png} & \includegraphics[width=.15\textwidth]{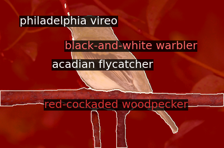} & \includegraphics[width=.15\textwidth]{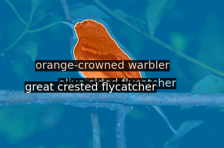} & \includegraphics[width=.15\textwidth]{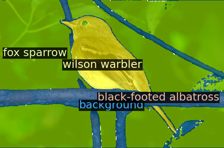} \\
\rotatebox[origin=l]{90}{CWFID} & \includegraphics[width=.15\textwidth]{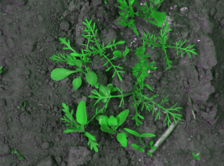} & \includegraphics[width=.15\textwidth]{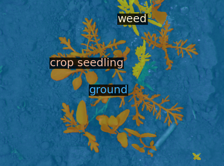} & \includegraphics[width=.15\textwidth]{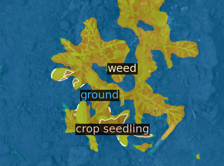} & \includegraphics[width=.15\textwidth]{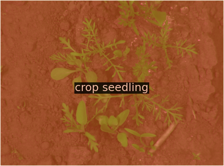} & \includegraphics[width=.15\textwidth]{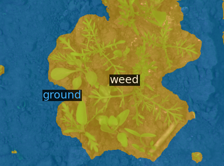} & \includegraphics[width=.15\textwidth]{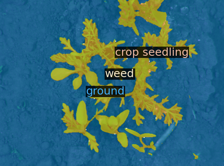} \\
\end{tabular}

    \caption{Predictions for datasets from medical sciences, engineering, and agriculture and biology.}
    \label{fig:predictions2}
\end{figure}

An example of each dataset with predictions from the four large models is presented in Figure~\ref{fig:predictions1} and~\ref{fig:predictions2}. CAT-Seg-L has visually the best predictions, which is in line with the quantitative results. The mask-based approaches SAN-L and OVSeg-L tend to segment very large areas with one class, e.g., in MHP v1, CryoNuSeg, and CWFID. Sometimes, they also fail to recognize the background as visualized in SUIM and CUB-200. This can happen when masks of the background include the predicted class itself. The prediction quality from Grounded-SAM-L varies the most. E.g., the model has a good prediction for UAVid but insufficient predictions for all other earth monitoring datasets.

\clearpage
\newpage

\bibliographystyle{apalike} 
\bibliography{references.bib}